\documentclass{article}

\PassOptionsToPackage{numbers,compress,sort&compress}{natbib}
\usepackage[main,final]{neurips_2026}

\usepackage[utf8]{inputenc}
\usepackage[T1]{fontenc}
\usepackage{url}
\usepackage{booktabs}
\usepackage{amsfonts}
\usepackage{amsmath}
\usepackage{amssymb}
\usepackage{nicefrac}
\usepackage{microtype}
\usepackage[table]{xcolor}
\usepackage{colortbl}
\usepackage{graphicx}
\usepackage{multirow}
\usepackage{xspace}
\usepackage{algorithm}
\usepackage{algpseudocode}
\usepackage{tikz}
\usetikzlibrary{arrows.meta,positioning}
\usepackage{hyperref}
\usepackage{subcaption}
\makeatletter
\renewenvironment{table}{\@float{table}}{\end@float}
\makeatother
\usepackage[breakable]{tcolorbox}
\usepackage{needspace}
\definecolor{mydarkgreen}{HTML}{1B7F3A}
\newcommand{\method}{\textsc{PROVE}\xspace}
\newcommand{\Mutator}{Mutator\xspace}
\newcommand{\Repairer}{Repairer\xspace}

\usepackage{multirow}
\usepackage{makecell}

\title{Perturb, Repair, Verify: Self-Play Vision-Language Verifiers for Compositional Understanding}

\author{%
  Zijun Huang\textsuperscript{1}\qquad
  Yingjun Du\textsuperscript{2}\thanks{Corresponding author.}\qquad
  Cees Snoek\textsuperscript{2}\\[4pt]
  \textsuperscript{1}University of Southern California \qquad
  \textsuperscript{2}VisLab, University of Amsterdam
}

\begin{document}

\maketitle

\begin{abstract}
Vision-language verifiers judge whether an image faithfully matches a textual description, yet existing training pipelines typically rely on static supervision from stronger external judges or curated hard-negative data. Such supervision requires external generation or annotation, is teacher-bounded, and is difficult to adapt as model failures evolve. We introduce \method (\textbf{P}erturb-and-\textbf{R}epair \textbf{O}ptimization for \textbf{V}ision-Language \textbf{E}valuation), a self-play framework that trains verifiers through a closed perturb--verify--repair--verify loop. Starting from a known-correct image-caption pair, a \Mutator introduces a typed perturbation over objects, attributes, counts, or relations. The verifier then judges the perturbed caption and provides textual feedback, which a \Repairer uses to recover a faithful caption for re-verification. These two verification rounds produce structured rewards for informative mismatches and visually grounded repairs. Since perturbation, verification, and repair are role-conditioned behaviors of the same evolving multimodal backbone, updating the shared parameters also improves the verifier itself. Experiments show that \method improves fine-grained compositional verification across model scales and families and strengthens visual-detail robustness on the reference backbone, with limited, benchmark-dependent changes on broader multimodal tasks.
\end{abstract}

\section{Introduction}
\label{sec:intro}

\begin{figure*}[t]
    \centering
    \begin{subfigure}[t]{0.32\textwidth}
        \centering
        \includegraphics[width=\linewidth]{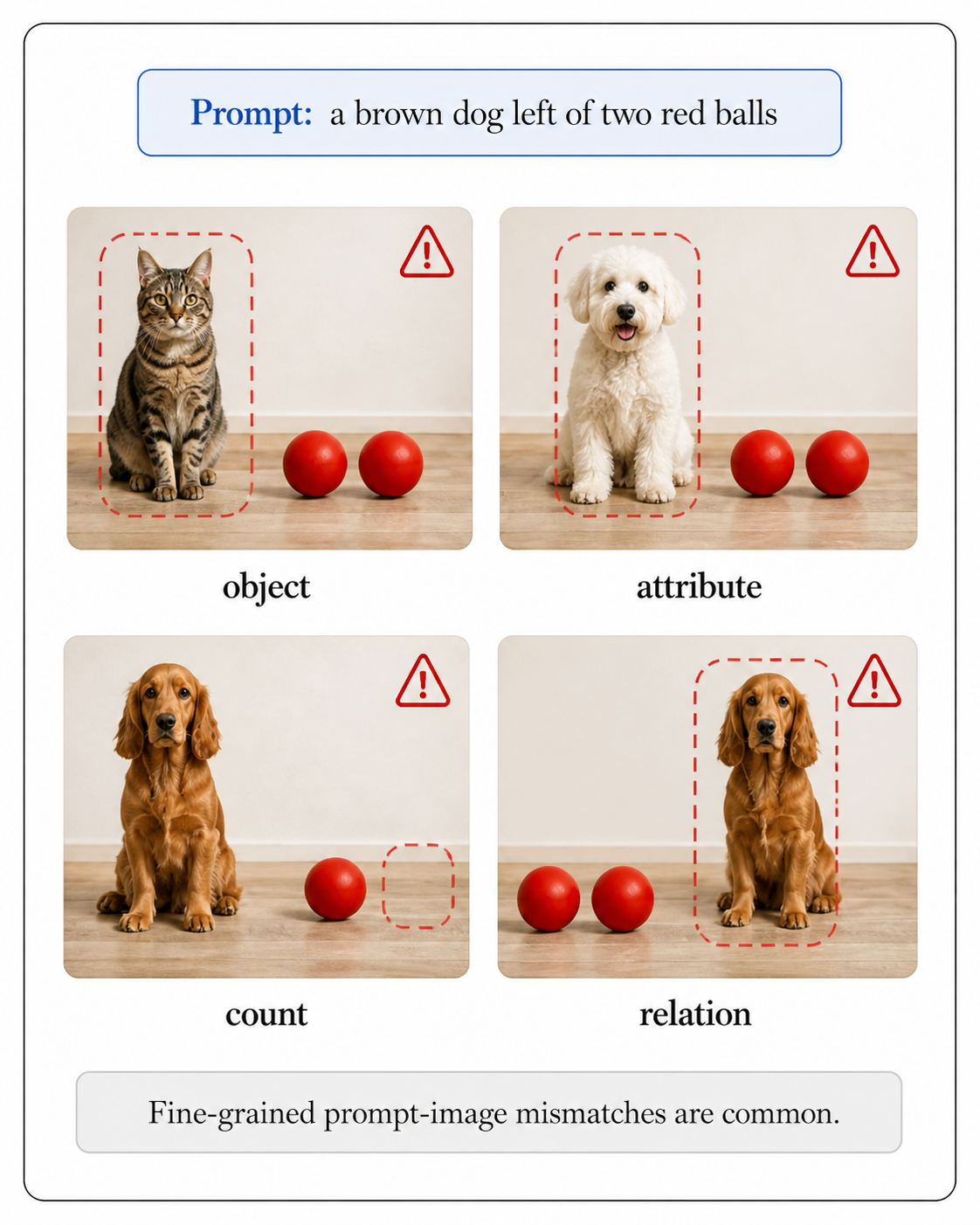}
        \caption{}
        \label{fig:motivation_failures}
    \end{subfigure}
    \hfill
    \begin{subfigure}[t]{0.32\textwidth}
        \centering
        \includegraphics[width=\linewidth]{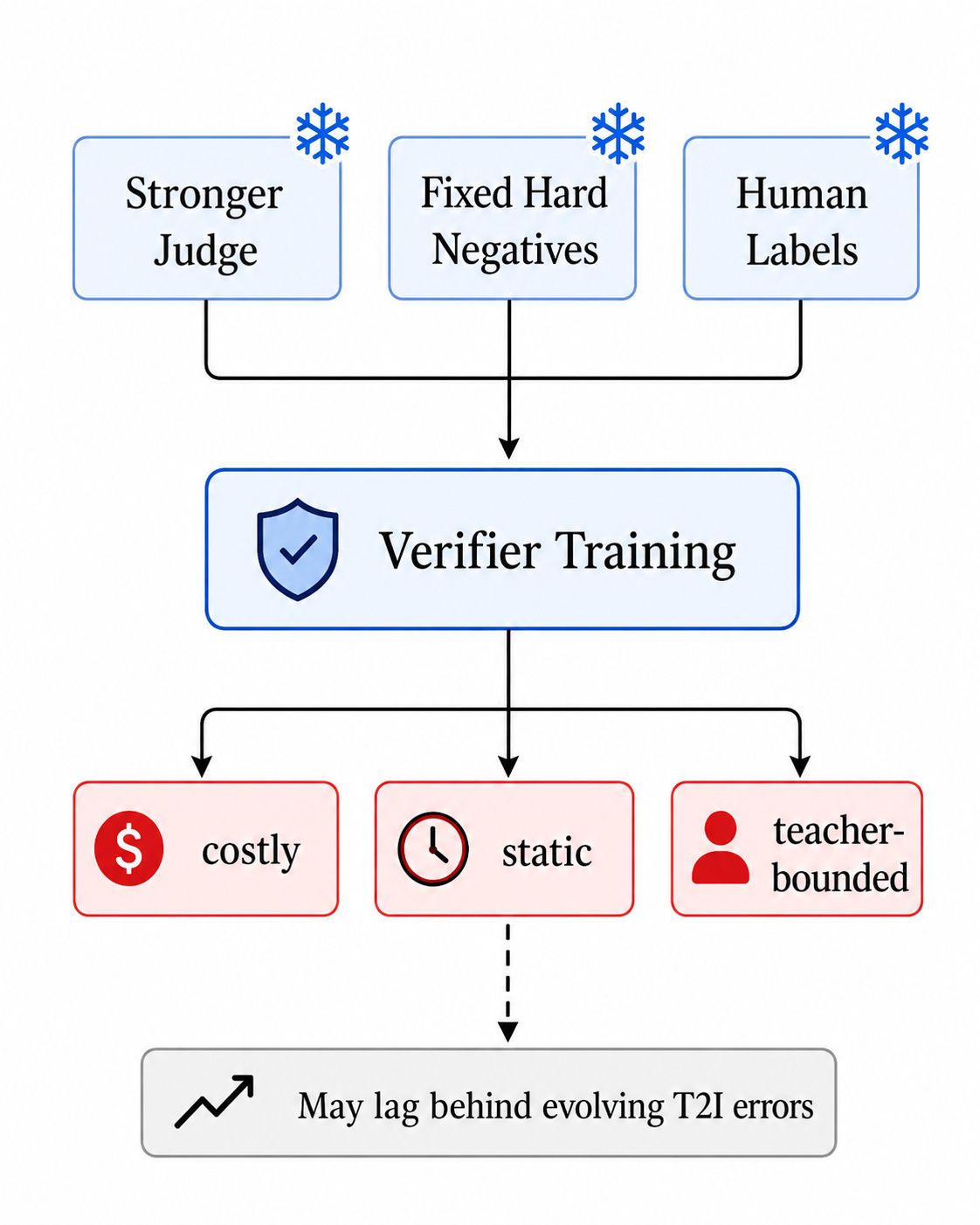}
        \caption{}
        \label{fig:motivation_static}
    \end{subfigure}
    \hfill
    \begin{subfigure}[t]{0.32\textwidth}
        \centering
        \includegraphics[width=\linewidth]{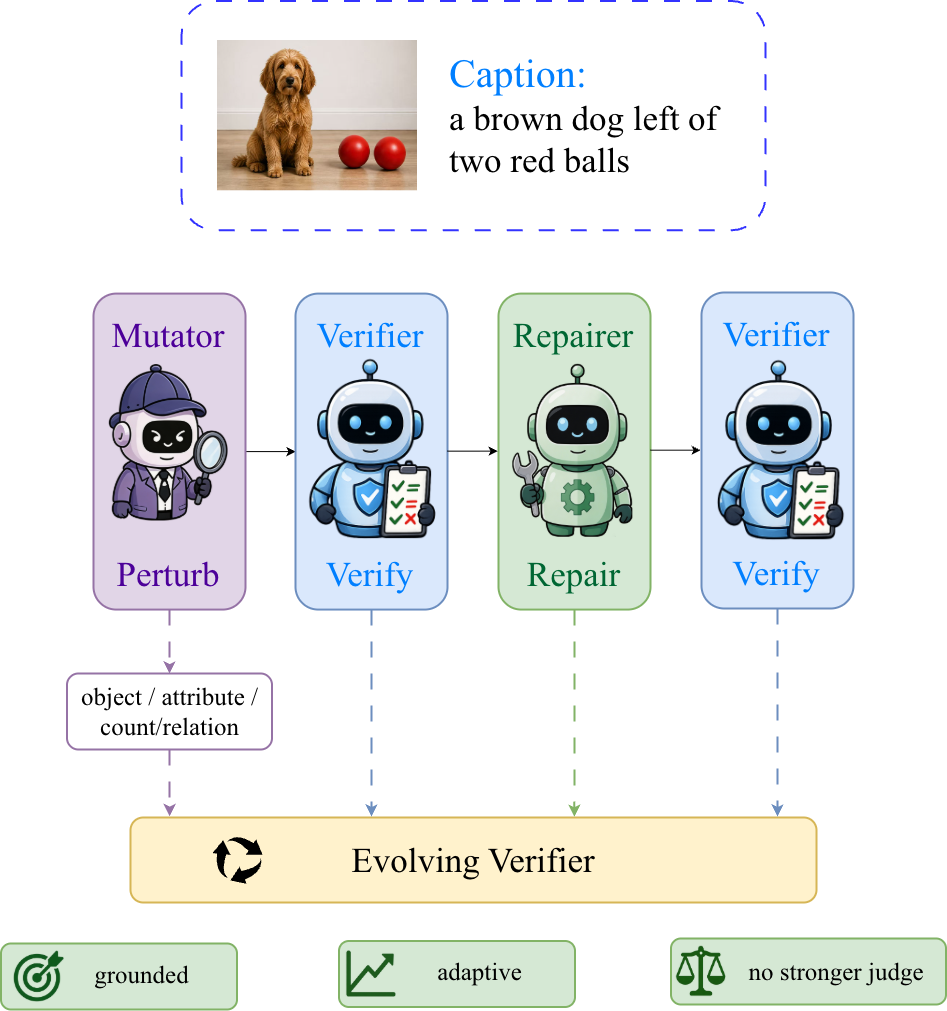}
        \caption{}
        \label{fig:motivation_prove}
    \end{subfigure}
\caption{
\textbf{Motivation of \method.}
(a) Fine-grained image-text verification requires detecting localized object, attribute, count, and relation mismatches.
(b) Existing verifier training relies on static external supervision, while (c) \method creates adaptive supervision through a perturb--verify--repair--verify self-play loop.
}
    \label{fig:intro_motivation}
    \vspace{-3mm}
\end{figure*}

Vision-language models have made rapid progress in image-text understanding, yet they still often miss subtle but consequential mismatches between an image and a textual description. A caption may differ from the image by a substituted object, a misbound attribute, an incorrect count, or a reversed relation, while still retaining high global semantic similarity. As illustrated in Fig.~\ref{fig:intro_motivation}\subref{fig:motivation_failures}, these errors lie along fine-grained visual-semantic axes that are difficult to capture with coarse image-text scores. This limitation is especially important because VLMs are increasingly used as \emph{verifiers}: given an image and a description, the verifier judges whether the description faithfully matches the image. Such verifiers support VLM-as-judge evaluation~\citep{zheng2023llmjudge}, reward modeling for RLHF~\citep{ouyang2022instructgpt} and RLAIF~\citep{lee2023rlaif}, generated-image filtering, and hallucination detection~\citep{li2023pope,guan2024hallusionbench}. A central question is therefore how to train verifiers that recognize localized image-text misalignment, rather than merely score global semantic compatibility.

The dominant recipe imports supervision from outside the target model, as summarized in Fig.~\ref{fig:intro_motivation}\subref{fig:motivation_static}. A stronger reference judge, typically a larger VLM or a curated process reward model, scores candidate image-text pairs, and the target verifier is trained to imitate or optimize against that score~\citep{zhang2025omniverifier,sun2023llavarlhf,lightman2024letsverify}. Alternatively, training relies on human-labeled or synthetically generated hard negatives~\citep{li2025sparcl,parascandolo2025cogt,jia2025visualleap}. Although effective, both routes provide \emph{static} supervision. The student is bounded by the perceptual distinctions exposed by the external judge or negative set; the supervision may lag behind evolving model failures; and extending verification to new axes such as counting, relations, or attribute binding often requires new annotation or generation pipelines. The core challenge is to obtain image-text mismatches that are simultaneously \emph{grounded} in the image, \emph{hard} enough to train fine-grained discrimination, and \emph{gradable} without relying on a stronger external judge.

We introduce \method (\textbf{P}erturb-and-\textbf{R}epair \textbf{O}ptimization for \textbf{V}ision-Language \textbf{E}valuation), a self-play framework that trains a verifier through a closed perturb--verify--repair--verify loop (Fig.~\ref{fig:intro_motivation}\subref{fig:motivation_prove}). Starting from a known-correct image-caption pair, a \Mutator creates a minimally perturbed caption by editing one visual-semantic axis, such as object identity, attribute, count, or relation. The verifier then judges the perturbed caption and provides feedback about the mismatch. Conditioned on this feedback, a \Repairer attempts to recover a faithful caption, which is verified again. Crucially, perturbation, verification, and repair are not separate models: they are three role-conditioned behaviors of the same evolving VLM. The two verification rounds produce rewards for informative perturbations and visually grounded repairs, while shared-parameter updates also improve the verifier used at evaluation time.

This formulation is well matched to fine-grained image-text verification. The \Mutator does not generate arbitrary text corruptions; it creates controlled, localized mismatches whose semantic axes mirror common compositional failures. The verifier must diagnose the mismatch before repair and recognize the corrected caption after repair, while the \Repairer must localize the error and infer the visually faithful description. In this way, \method converts ordinary clean image-caption pairs into adaptive hard negatives and repair targets. Compared with external-judge distillation, the training signal is not bounded by a stronger frozen oracle. Compared with fixed hard-negative training, the mismatch distribution evolves with the current verifier.

We evaluate \method on a Qwen3.5-2B backbone across fine-grained compositional alignment benchmarks and broader vision-language robustness benchmarks. \method consistently improves the same backbone on compositional verification, with the largest gains appearing on axes aligned with the perturbation space, such as attribute and relation understanding. It also improves visual-detail sensitivity and maintains hallucination robustness, while broader capability changes are limited and benchmark-dependent. Training diagnostics further show that perturbation and repair co-adapt through verifier feedback. These results suggest that vision-language verifiers can be strengthened through verifier-guided self-play rather than static judge distillation.

Matched supervision baselines, independent verifier audits, and additional Qwen3.5-4B and InternVL3-2B experiments further support this teacher-free adaptive alternative. The present study establishes caption-based verification and benchmark transfer; task-specific self-play for VQA, grounding, and referring expressions remains future work.

\section{Related Work}
\label{sec:related}

\noindent\textbf{Vision-language verifiers for image-text understanding.}
Evaluating whether an image faithfully matches a textual description is a central problem in vision-language understanding. Coarse image-text similarity metrics are often insufficient, because they can miss fine-grained errors in object identity, attribute binding, counting, and spatial relations. Recent work therefore uses vision-language models as verifiers or judges for multimodal evaluation, reward modeling, generated-image filtering, and hallucination detection~\citep{zheng2023llmjudge,li2023pope,guan2024hallusionbench,du2024ipo}. These verifiers are attractive because they can provide semantic judgments beyond fixed embedding similarities, but their reliability depends on whether they can detect subtle image-text mismatches. \method targets this capability directly by training the verifier through controlled perturbation, verification, and repair over the same compositional axes that commonly challenge image-text understanding.

\noindent\textbf{Training verifiers with static external supervision.}
A common way to train multimodal verifiers is to import supervision from outside the target model. For example, stronger VLMs or curated reward models can score candidate image-text pairs, after which the target model is optimized by preference learning or reward-based training~\citep{zhang2025omniverifier,sun2023llavarlhf,li2024vlfeedback,lightman2024letsverify,sun2026less,sun2026regionreasoner}. Another line constructs fixed hard-negative datasets or synthetic preference pairs, then trains the verifier to distinguish faithful captions from controlled corruptions~\citep{lin2025selfimprovingjudge}. These approaches provide useful supervision, but the signal is static: it is bounded by the external judge, fixed negative generator, or annotation pipeline. As models and evaluation distributions evolve, their failure modes may shift away from this supervision. In contrast, \method does not distill a stronger frozen judge or rely on a fixed hard-negative set. It uses the current verifier itself inside a perturb--verify--repair--verify loop, so the training signal changes with the evolving model.

\noindent\textbf{Self-play and self-evolving multimodal learning.}
Self-play has been widely used to improve agents and reasoning systems, from game-playing methods such as AlphaZero~\citep{silver2018alphazero} to debate and recent self-evolving LLMs~\citep{irving2018debate,huang2026rzero,zhao2025absolutezero}. More recent asymmetric self-play methods train one behavior to propose tasks and another to solve them, creating an adaptive curriculum without a fixed external teacher~\citep{liu2025spice}. In multimodal learning, self-evolving VLMs have also been explored through questioner--reasoner interactions and pseudo-labeling from model samples~\citep{he2025visplay}. However, consensus- or uncertainty-based supervision can still reinforce confident mistakes, especially when the model fails to perceive the relevant visual evidence. \method differs in both task and interaction structure: it trains a vision-language verifier rather than a general reasoner, and it grounds self-play in clean image-caption pairs. The \Mutator creates typed image-text mismatches, the verifier diagnoses them, and the \Repairer attempts to correct them; all three are role-conditioned behaviors of the same evolving MLLM.

\noindent\textbf{Compositional image-text alignment.}
Vision-language models often struggle with compositional understanding despite strong global image-text alignment. Benchmarks such as Winoground~\citep{thrush2022winoground}, ARO~\citep{yuksekgonul2023aro}, SugarCrepe~\citep{hsieh2023sugarcrepe}, and VL-CheckList~\citep{zhao2022vlchecklist} test whether models can distinguish captions that differ in objects, attributes, relations, or word order. Prior methods improve this ability with hard-negative contrastive learning or synthetic data~\citep{yuksekgonul2023aro,doveh2023dac}. \method is complementary: it generates controlled perturbations online and uses verifier-guided repair to create adaptive supervision for fine-grained image-text alignment.

\section{Methods}
\label{sec:method}

\subsection{Overview}

\begin{figure}[t]
  \centering
    \includegraphics[width=\textwidth,clip]{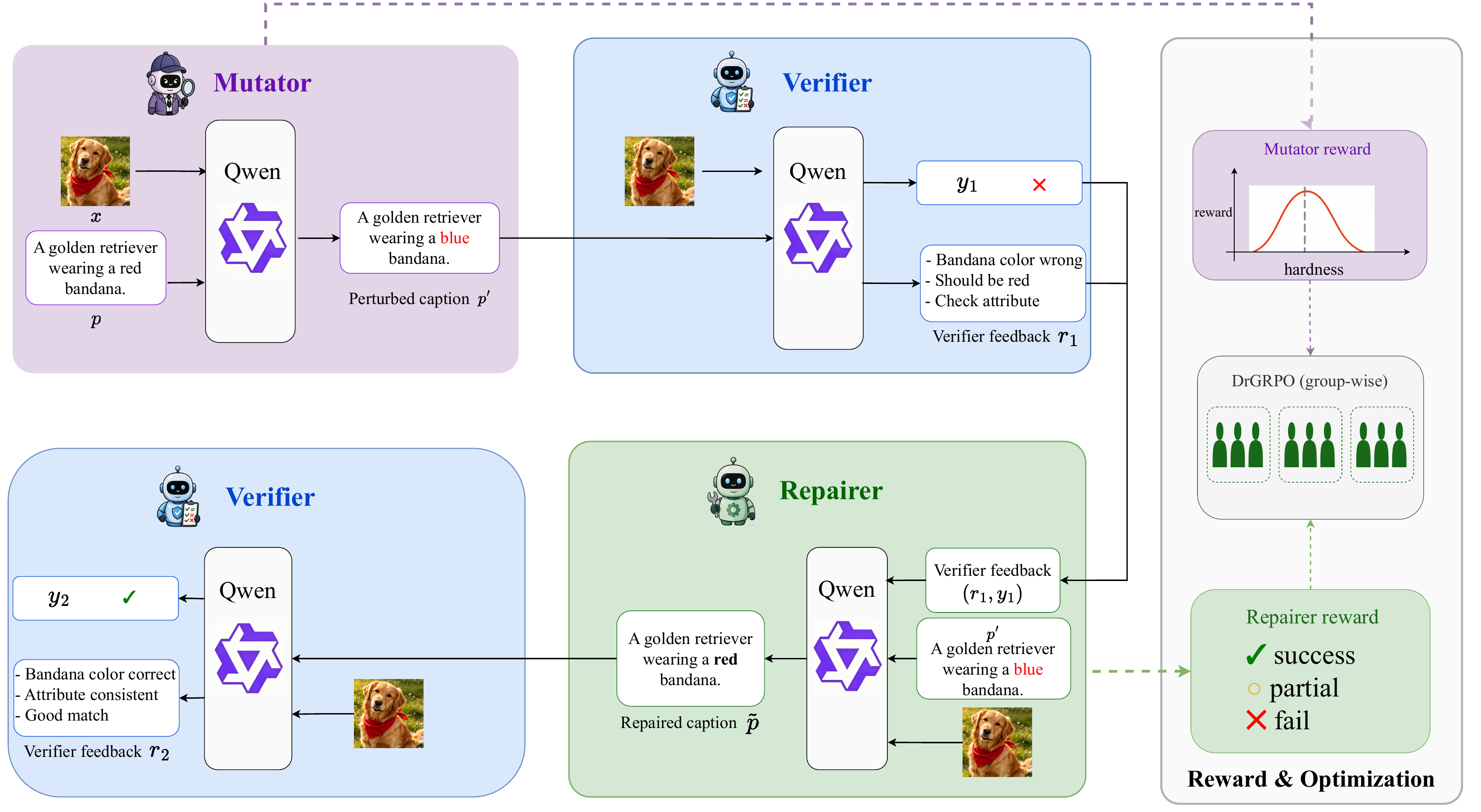}
\caption{
\textbf{Overview of \method.}
A single Qwen backbone is reused under three role prompts: \Mutator, Verifier, and \Repairer.
Given a clean image-caption pair, the \Mutator produces a localized caption perturbation, e.g., changing an attribute from ``red'' to ``blue'', to create a hard image-text mismatch.
The Verifier first judges the perturbed caption and returns a decision-feedback pair $(y_1,r_1)$ that diagnoses the mismatch.
Conditioned on this feedback, the \Repairer generates a revised caption, which is then checked by the same Verifier to obtain $(y_2,r_2)$.
Rewards from the two verification rounds train the shared model with DrGRPO, so perturbation, verification, and repair co-evolve within one multimodal backbone.
}
  \label{fig:overview}
\end{figure}

\method trains a vision-language verifier through a closed perturb--verify--repair--verify loop.
Given a clean image-caption pair, a single multimodal policy $\pi_\theta$ is reused under three role prompts: a \Mutator that perturbs the caption, a Verifier that judges image-caption faithfulness, and a \Repairer that corrects the caption using verifier feedback.
These roles are not separate models; they are three behaviors of the same parameter-shared backbone.
Thus, every policy update also changes the verifier used at evaluation time.
Figure~\ref{fig:overview} illustrates the overall loop.

The key idea is to turn ordinary positive image-caption pairs into adaptive verification episodes.
The \Mutator creates localized mismatches along object, attribute, count, or relation axes.
The Verifier diagnoses whether the perturbed caption still matches the image.
The \Repairer then attempts to recover a faithful caption from the image and verifier feedback.
This interaction produces grounded hard negatives and repair targets without a stronger external judge or a fixed hard-negative dataset.

\subsection{Perturb--Verify--Repair--Verify Self-Play}
\label{sec:selfplay}

Let $\mathcal{D}=\{(x_i,p_i)\}_{i=1}^{N}$ be a set of image-caption pairs, where $p_i$ faithfully describes image $x_i$.
For a clean pair $(x,p)$, \method constructs the following self-play episode:
\begin{equation}
\label{eq:prove_episode}
(x,p)
\;\xrightarrow{\text{\Mutator}}\;
p'
\;\xrightarrow{\text{Verifier}}\;
(y_1,r_1)
\;\xrightarrow{\text{\Repairer}}\;
\tilde p
\;\xrightarrow{\text{Verifier}}\;
(y_2,r_2).
\end{equation}
Here $p'$ is a perturbed caption, $(y_1,r_1)$ is the first verifier decision and explanation, $\tilde p$ is the repaired caption, and $(y_2,r_2)$ is the second verifier decision and explanation.
The binary label $y\in\{0,1\}$ indicates whether the caption is judged faithful to the image, and $r$ is natural-language feedback.

All stages are sampled from the same policy under different role prompts:
\begin{equation}
\label{eq:role_policy}
\pi_\theta(\cdot \mid x,q,z),
\qquad
z\in\{\textsc{Mutator},\textsc{Verifier},\textsc{Repairer}\},
\end{equation}
where $q$ denotes the role-specific text context.
This makes the verifier part of the evolving policy, rather than a frozen reward model.

\paragraph{Caption perturbation.}
The \Mutator produces a localized edit
\begin{equation}
\label{eq:mutation_action}
a_M=(\tau,t,\delta),
\end{equation}
where $\tau$ is the edit type, $t$ is the target span, and $\delta$ specifies the replacement.
We use four edit types:
\begin{equation}
\label{eq:perturb_types}
\mathcal{T}=\{\textsf{object},\textsf{attribute},\textsf{count},\textsf{relation}\}.
\end{equation}
A deterministic editor applies the edit to obtain the perturbed caption:
\begin{equation}
\label{eq:perturbed_caption}
p'=E(p,a_M).
\end{equation}
Restricting the \Mutator to typed local edits prevents arbitrary rewriting and keeps the negative caption close to the original image-caption pair.
Invalid edit types, unmatched or ambiguous target spans, unchanged replacements, and non-deterministic edits are filtered; Appendix~\ref{app:interfaces} describes the role interfaces and editor constraints.

\paragraph{Verification and repair.}
The Verifier first judges the perturbed caption:
\begin{equation}
\label{eq:first_verify}
(y_1,r_1)\sim \pi_\theta(\cdot \mid x,p',\textsc{Verifier}).
\end{equation}
The feedback $r_1$ identifies the perceived mismatch and is passed to the \Repairer, which generates a corrected caption:
\begin{equation}
\label{eq:repair_caption}
\tilde p\sim \pi_\theta(\cdot \mid x,p',y_1,r_1,\textsc{Repairer}).
\end{equation}
The same Verifier then evaluates the repaired caption:
\begin{equation}
\label{eq:second_verify}
(y_2,r_2)\sim \pi_\theta(\cdot \mid x,\tilde p,\textsc{Verifier}).
\end{equation}
The first verification tests whether the model detects the perturbation; the second tests whether the repair restores faithfulness.
Together, they turn verification into an operational correction task.

\subsection{Verifier-Guided Rewards}
\label{sec:reward}

The two verifier calls provide the supervision for self-play.
The \Mutator should produce perturbations that are informative, grounded, and not overly destructive.
The \Repairer should recover a faithful caption while making minimal changes.

\paragraph{Mutator reward.}
We define the \Mutator reward as
\begin{equation}
\label{eq:mutator_reward_concept}
r_M
=
R_{\mathrm{expose}}(y_1)
+
R_{\mathrm{hard}}(y_2)
-
\lambda_M C_M(a_M),
\end{equation}
where $R_{\mathrm{expose}}$ rewards perturbations that expose verifier blind spots, $R_{\mathrm{hard}}$ rewards perturbations that remain difficult after repair, and $C_M$ penalizes overly wide edits. Invalid actions receive a fixed penalty and do not enter $C_M$ (Appendix~\ref{app:interfaces}).
We instantiate it as
\begin{equation}
\label{eq:mutator_reward_inst}
r_M
=
\lambda_1\mathbb{1}[y_1=1]
+
\lambda_2\mathbb{1}[y_2=0]
-
\lambda_3 C_M(a_M).
\end{equation}

\paragraph{Repairer reward.}
The \Repairer reward encourages a successful and minimal correction:
\begin{equation}
\label{eq:repairer_reward_concept}
r_R
=
R_{\mathrm{accept}}(y_2)
+
R_{\mathrm{flip}}(y_1,y_2)
-
\lambda_R C_R(p',\tilde p),
\end{equation}
where $R_{\mathrm{accept}}$ rewards repaired captions accepted by the verifier, $R_{\mathrm{flip}}$ rewards changing the verifier judgment, and $C_R$ penalizes unnecessary rewriting.
We use
\begin{equation}
\label{eq:repairer_reward_inst}
r_R
=
\beta_1\mathbb{1}[y_2=1]
+
\beta_2\mathbb{1}[y_1\ne y_2]
-
\beta_3 C_R(p',\tilde p).
\end{equation}

Although rewards are assigned to perturbation and repair trajectories, the Verifier is not a frozen judge.
Because all roles share $\theta$, updates from these rewards also update the verifier behavior itself.
This coupling is what allows \method to improve the verifier without distilling a stronger external model.

\subsection{Role-Conditioned Policy Optimization}
\label{sec:optimization}

We optimize the shared policy with DrGRPO~\cite{liu2025understanding}.
For each source pair, the model samples multiple self-play trajectories and computes verifier-guided rewards.
Let $r_i$ be the reward of trajectory $i$, and let $G(i)$ be the group of trajectories sampled from the same source context.
Following DrGRPO, we use mean-centered advantages without standard-deviation normalization:
\begin{equation}
\label{eq:grpo_advantage}
\hat A_i
=
r_i
-
\frac{1}{|G(i)|}
\sum_{j\in G(i)} r_j .
\end{equation}
This preserves the reward scale and avoids unstable rescaling when within-group variance is small.

The clipped objective over generated tokens is
\begin{equation}
\label{eq:grpo_objective}
\mathcal{L}(\theta)
=
\mathbb{E}_{i,t}
\left[
\min\left(
\rho_{i,t}(\theta)\hat A_i,
\operatorname{clip}\!\left(\rho_{i,t}(\theta),1-\epsilon^{-},1+\epsilon^{+}\right)\hat A_i
\right)
\right],
\end{equation}
where
\begin{equation}
\label{eq:ratio_new}
\rho_{i,t}(\theta)
=
\frac{
\pi_\theta(a_{i,t}\mid q_i,a_{i,<t})
}{
\pi_{\theta_{\mathrm{old}}}(a_{i,t}\mid q_i,a_{i,<t})
}.
\end{equation}
Here $q_i$ is the role-conditioned multimodal prompt and $a_{i,t}$ is the generated token.
Equation~\ref{eq:grpo_objective} is a maximization objective; equivalently, its negative is minimized, with KL regularization specified in the experimental setup.
At test time, we use the Verifier prompt of the evolved shared model to judge image-caption faithfulness.

\section{Experiments}
\label{sec:experiments}
\subsection{Experimental Setup}

\noindent\textbf{Backbone and training framework.}
We instantiate \method with Qwen3.5-2B~\cite{qwenteam2026qwen35} as the shared multimodal backbone. 
The same parameters are reused across the \Mutator, Verifier, and \Repairer roles, with role identity specified only by the prompt template. 
Thus, perturbation, verification, and repair are optimized as role-conditioned behaviors of one evolving MLLM rather than as separate models. 
At evaluation time, we use the Verifier role of the evolved shared model to judge image-caption faithfulness.

\noindent\textbf{Training data and self-play construction.}
We train on a 48k-example subset of ShareGPT-4o-Image-style image-caption pairs~\cite{chen2025sharegpt4oimage}, with a held-out validation split of 5k examples. 
Each self-play episode starts from a known-correct image-caption pair. 
The \Mutator introduces a typed perturbation along one of four visual-semantic axes: object, attribute, count, or relation. 
The Verifier then judges the perturbed caption and produces a decision-feedback pair. 
Conditioned on this feedback, the \Repairer generates a repaired caption, which is judged again by the same Verifier role. 
The two verification rounds provide verifier-guided rewards for self-play optimization. 
No stronger external judge, human-labeled hard negatives, or additional hard-negative annotations are used during training.

\noindent\textbf{Optimization and compute.}
We implement \method using the VeRL/HybridFlow training framework~\cite{sheng2025hybridflow} and optimize the shared policy with DrGRPO~\cite{liu2025understanding}, a group-relative policy optimization variant that removes the standard-deviation normalization used in vanilla GRPO~\cite{shao2024deepseekmath}. 
Unless otherwise specified, we use rollout size $n=3$, learning rate $5\times10^{-7}$, global batch size 48, KL coefficient $0.005$, and PPO clip range $[0.20,0.22]$. 
All main training runs are conducted on a single node with 8 NVIDIA A100 GPUs using BF16 mixed precision. 
We report the best validation checkpoint in the main results. 
Training hyperparameters, role interfaces, and additional experimental results are provided in the appendix.
The reference run takes 11 hours on 8 A100 GPUs (88 GPU-hours); Appendix~\ref{app:compute} compares complete pipeline costs.
We will release the training and evaluation code to facilitate reproducibility.

\noindent\textbf{Evaluation protocol.}
Our main evaluation focuses on fine-grained image-text understanding, where a verifier must distinguish captions that differ by localized semantic changes. 
Table~\ref{tab:cross_model} compares \method with representative compositional understanding methods on four benchmarks shared with \cite{parascandolo2025cogt}: ARO~\cite{yuksekgonul2023aro}, VL-CheckList~\cite{zhao2022vlchecklist}, ColorSwap~\cite{burapacheep2024colorswap}, and FG-OVD~\cite{bianchi2024fgovd}. 
These benchmarks cover relation and attribute reasoning, object-level grounding, color-word order sensitivity, and fine-grained open-vocabulary recognition. 
For each baseline, we report the values published in the original paper whenever available; when a benchmark was not reported but public code is available, we use our reproduction under the same evaluation protocol.

Beyond the core compositional suites, we evaluate how the learned verifier transfers to broader vision-language understanding.
We include Winoground~\cite{thrush2022winoground} to test role binding under hard image-text contrasts, MMVP~\cite{tong2024mmvp} to assess sensitivity to fine visual details, and MMBench-EN~\cite{liu2024mmbench} to measure general multimodal perception and reasoning. 
For all two-alternative forced-choice tasks, we randomize option order for each example to reduce positional bias. 
Unless otherwise stated, all benchmarks are evaluated using their official protocols and metrics.
Comp. Avg. averages the four core benchmark groups in Table~\ref{tab:cross_model}; Broader Avg. averages Winoground GS, MMVP pair accuracy, and MMBench in Table~\ref{tab:additional_benchmarks}. Winoground GS and the appendix 2-AFC accuracy are distinct metrics. External evaluation labels are not used for training, rewards, mutation generation, validation, or checkpoint selection.

\subsection{Main Results}

\begin{table*}[t]
\centering
\caption{
\textbf{Comparison with compositional understanding methods.}
We report all available sub-metrics on ARO, VL-CheckList, ColorSwap, and FG-OVD. ``--'' indicates that the benchmark was not evaluated.
Best and second-best results are shown in \textbf{bold} and \underline{underline}.
\method achieves the best overall average among methods with complete benchmark coverage and consistently improves over the Qwen3.5-2B backbone across all benchmark groups.
}
\label{tab:cross_model}
\definecolor{oursgray}{gray}{0.92}
\scalebox{0.8}{
\begin{tabular}{@{}l|ccc|cccc|c|c|c@{}}
\toprule
\textbf{Model} 
& \multicolumn{3}{c|}{\textbf{ARO}} 
& \multicolumn{4}{c|}{\textbf{VL-CheckList}} 
& \textbf{ColorSwap} 
& \textbf{FG-OVD} 
& \textbf{Overall} \\
\cmidrule(lr){2-4}
\cmidrule(lr){5-8}
\cmidrule(lr){9-9}
\cmidrule(lr){10-10}
\cmidrule(l){11-11}
& Rel & Attr & Avg 
& Attr & Obj & Rel & Avg 
& ITT & Avg & Avg \\
\midrule
\rowcolor{gray!12}
\multicolumn{11}{@{}l}{\textit{Contextual comparison: discriminative VLMs}} \\
CLIP~\cite{radford2021clip}         
& 59.00 & 62.00 & 60.50 
& 67.93 & 82.83 & 64.19 & 71.65 
& 35.67 & 47.33 & 53.79 \\
NegCLIP~\cite{yuksekgonul2023aro}   
& 81.00 & 71.00 & 76.00 
& 72.24 & 87.00 & 71.39 & 76.87 
& 35.67 & 41.69 & 57.56 \\
GNM~\cite{sahin2024gnm}             
& 65.00 & 65.00 & 65.00 
& 70.15 & 85.91 & 64.10 & 73.38 
& 13.00 & 38.79 & 47.54 \\
PANeg~\cite{buettner2024plausible}  
& 65.07 & 67.94 & 66.51 
& 76.51 & 88.13 & 69.90 & 78.17 
& 17.67 & 44.98 & 51.83 \\
DAC-SAM~\cite{doveh2023dac}         
& 77.16 & 70.50 & 73.83 
& 75.80 & 88.50 & \underline{89.80} & \underline{84.70} 
& 16.33 & 48.36 & 55.81 \\
DAC-LLM~\cite{doveh2023dac}         
& 81.28 & 73.91 & 77.60 
& 77.30 & 87.30 & 86.40 & 83.66 
& 18.33 & 49.60 & 57.30 \\
XVLM~\cite{zeng2022xvlm}            
& 73.40 & 86.80 & 80.10 
& 75.10 & 85.80 & 70.40 & 76.50 
& -- & -- & -- \\
\midrule
\rowcolor{gray!12}
\multicolumn{11}{@{}l}{\textit{Contextual comparison: generative VLMs}} \\
InstructBLIP~\cite{dai2023instructblip} 
& 69.20 & 50.83 & 60.02 
& 56.37 & 80.33 & 53.34 & 63.35 
& 40.33 & 26.80 & 47.63 \\
MiniGPT-4~\cite{zhu2023minigpt4}    
& 46.90 & 55.70 & 51.30 
& 71.30 & 84.20 & -- & -- 
& -- & -- & -- \\
BLIP2-VGPS~\cite{lin2024visualgptscore} 
& \textbf{90.70} & \textbf{94.30} & \textbf{92.50} 
& 73.90 & 90.10 & \textbf{89.90} & 84.63 
& -- & -- & -- \\
CapPa~\cite{tschannen2023cap}       
& \underline{86.70} & 85.70 & 86.20 
& -- & -- & -- & -- 
& -- & -- & -- \\
LLaVA-1.5-7B~\cite{liu2024llava15}  
& 65.08 & 57.40 & 61.24 
& 69.49 & -- & 73.50 & -- 
& -- & -- & -- \\
InternVL2-8B~\cite{chen2024internvl2} 
& 83.42 & \underline{93.74} & \underline{88.58} 
& \textbf{84.81} & -- & 87.10 & -- 
& -- & -- & -- \\
\midrule
\rowcolor{gray!12}
\multicolumn{11}{@{}l}{\textit{Matched comparison: shared-backbone verifier models}} \\
Qwen3.5-2B~\cite{qwenteam2026qwen35} 
& 69.68 & 79.48 & 74.58 
& 76.75 & \underline{91.20} & 82.85 & 83.60 
& \underline{93.67} & \underline{81.01} & \underline{83.22} \\
\rowcolor{blue!4}
\method (Qwen3.5-2B) 
& 76.92 & 86.74 & 81.83 
& \underline{81.46} & \textbf{94.14} & 86.95 & \textbf{87.52} 
& \textbf{95.83} & \textbf{82.23} & \textbf{86.85} \\
\bottomrule
\end{tabular}}%
\end{table*}

\begin{table}[t]
\centering
\caption{
\textbf{Comparison on broader vision-language benchmarks.}
\method achieves the best average and improves the Qwen3.5-2B backbone on all three benchmarks, indicating transfer beyond the core compositional suites. Broader task changes are benchmark-dependent (Appendix~\ref{app:seeds}).
}
\label{tab:additional_benchmarks}
\scalebox{1.}{
\begin{tabular}{@{}l|cccc@{}}
\toprule
\textbf{Model} & \textbf{Wino GS} & \textbf{MMVP Pair} & \textbf{MMBench} & \textbf{Avg} \\
\midrule
\rowcolor{gray!12}
\multicolumn{5}{@{}l}{\textit{Discriminative VLMs}} \\
CLIP~\cite{radford2021clip}        & 8.00  & 15.55 & --    & -- \\
NegCLIP~\cite{yuksekgonul2023aro}  & 8.00  & 26.67 & --    & -- \\
PANeg~\cite{buettner2024plausible} & 6.75  & 21.48 & --    & -- \\
DAC-SAM~\cite{doveh2023dac}        & 6.25  & 23.70 & --    & -- \\
DAC-LLM~\cite{doveh2023dac}        & 3.25  & 28.14 & --    & -- \\
\midrule
\rowcolor{gray!12}
\multicolumn{5}{@{}l}{\textit{Generative VLMs}} \\
InstructBLIP~\cite{dai2023instructblip} & 4.75  & 14.81 & --    & -- \\
LLaVA-1.5-7B~\cite{liu2024llava15}      & 27.00 & 24.70 & 64.30 & 38.67 \\
InternVL2-8B~\cite{chen2024internvl2}   & \underline{49.75} & 51.33 & \textbf{81.70} & \underline{60.93} \\
\midrule
\rowcolor{gray!12}
\multicolumn{5}{@{}l}{\textit{Shared-backbone verifier models}} \\
Qwen3.5-2B~\cite{qwenteam2026qwen35} 
& 49.00 & \underline{56.00} & 71.31 & 58.77 \\
\rowcolor{blue!4}
\method (Qwen3.5-2B) 
& \textbf{50.75} & \textbf{60.00} & \underline{72.34} & \textbf{61.03} \\
\bottomrule
\end{tabular}}
\end{table}

\noindent\textbf{Compositional understanding.}
Table~\ref{tab:cross_model} compares \method with representative compositional understanding methods on ARO, VL-CheckList, ColorSwap, and FG-OVD. 
Because these methods differ in backbone architecture, training data, objectives, and benchmark coverage, the cross-method comparison should be interpreted as a contextual reference rather than a fully controlled comparison. 
Our main controlled comparison is between Qwen3.5-2B and \method using the same backbone.
Under this controlled setting, \method improves the overall average from 83.22 to 86.85, with consistent gains across all four benchmark groups. 
The largest improvement appears on ARO, where the average score increases by $+7.25$, indicating stronger attribute and relation discrimination. 
\method also improves VL-CheckList, ColorSwap, and FG-OVD, suggesting that the learned verifier benefits not only the perturbation axes seen during self-play but also broader forms of localized image-text grounding. 
Compared with prior compositional methods, \method achieves the best overall average among complete rows in this table and remains competitive on individual benchmark groups, despite using a compact shared backbone and no external hard-negative dataset.
These results support the core design of \method. 
The \Mutator creates controlled mismatches over objects, attributes, counts, and relations; the Verifier learns to diagnose these mismatches; and the \Repairer learns to recover visually faithful captions. 
The resulting self-play signal strengthens fine-grained image-text verification without relying on a stronger external judge.

\noindent\textbf{Broader vision-language understanding benchmarks.}
Table~\ref{tab:additional_benchmarks} evaluates whether the learned verifier transfers beyond the main compositional suites. 
\method improves the Qwen3.5-2B backbone on all three benchmarks, raising the average from 58.77 to 61.03. 
The most notable gain is on MMVP, where pair accuracy improves by $+4.00$, suggesting improved sensitivity to fine visual details. 
Winoground also improves, indicating better role binding under hard image-text contrasts, while MMBench-EN shows a modest positive change, suggesting positive transfer to this general multimodal perception and reasoning benchmark.

Together, Tables~\ref{tab:cross_model} and~\ref{tab:additional_benchmarks} show that \method mainly strengthens the verifier on localized image-text discrimination, which matches the intended training signal of typed perturbation and verifier-guided repair. 
At the same time, the gains on MMVP and MMBench indicate positive transfer, while repeated-run results reveal small, task-dependent regressions on other broader benchmarks (Appendix~\ref{app:seeds}).

\begin{table}[t]
\centering
\caption{\textbf{Matched supervision and core loop ablations.} All training variants use Qwen3.5-2B, 48k source pairs, and the same target-model optimization budget and evaluation protocol. Complete ablations are in Appendix~\ref{app:controls}.}
\label{tab:controlled}
\small
\begin{tabular}{@{}lcc@{}}
\toprule
Method / variant & Comp. Avg. & Broader Avg. \\
\midrule
Qwen3.5-2B & 83.22 & 58.77 \\
Fixed typed negatives & 85.02 & 59.68 \\
External-judge distillation & 86.29 & 60.72 \\
Verifier self-training & 85.66 & 60.21 \\
\midrule
No \Repairer or second verification & 84.24 & 59.31 \\
\Repairer without verifier feedback & 85.17 & 59.84 \\
Frozen internal Verifier & 85.12 & 59.81 \\
\midrule
\rowcolor{blue!4}
\method & \textbf{86.85} & \textbf{61.03} \\
\bottomrule
\end{tabular}
\end{table}

\noindent\textbf{Matched supervision alternatives.}
Table~\ref{tab:controlled} compares static negatives generated once, judgments from a frozen larger VLM, and self-generated verifier labels without repair.
\method improves the compositional/broader averages over these baselines by $1.83/1.35$, $0.56/0.31$, and $1.19/0.82$ points, respectively.
These results support adaptive perturb--repair self-play as a teacher-free alternative; its advantage is adaptive supervision rather than uniformly lower compute or universal superiority over distillation.

\begin{figure}[t]
  \centering
  \includegraphics[width=\linewidth]{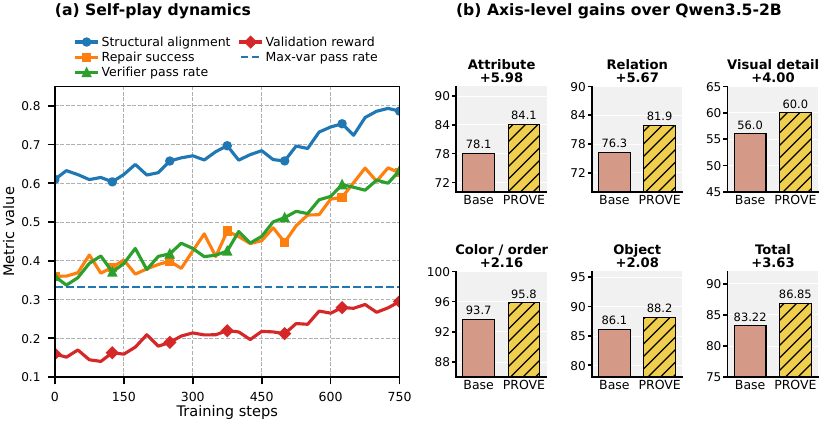}
\caption{
\textbf{Self-play dynamics and axis-level gains.}
(a) The \Repairer steadily improves in repair success and structural alignment, while the verifier pass rate rises as repairs catch up with the current \Mutator distribution; the dashed line marks the lower finite-group maximum-variance reference at $1/3$ (Appendix~\ref{app:interfaces}).
(b) Each panel compares Qwen3.5-2B with \method on an aggregated visual-semantic capability; Total is the overall average in Table~\ref{tab:cross_model}.
The largest gains appear on attribute and relation verification, matching the typed perturbation axes used during self-play.
}
  \label{fig:dynamics}
  \label{fig:axis_gains}
\end{figure}

\subsection{Analysis}
\label{sec:analysis}

\noindent\textbf{Self-Play Dynamics Analysis.} Figure~\ref{fig:dynamics}(a) analyzes how the perturb--verify--repair loop evolves during training.
The training dynamics show that the \Repairer steadily catches up with the perturbations produced by the \Mutator while the overall optimization remains stable. 
Repair success increases from $36.0\%$ to $62.7\%$, and structural alignment rises from $61.0\%$ to $78.6\%$, indicating that the \Repairer becomes increasingly capable of localizing \Mutator-induced mismatches and recovering visually faithful captions. 
At the same time, the verifier pass rate increases from $0.360$ to $0.633$ and remains above the lower finite-group maximum-variance reference point of $1/3$, suggesting that the current four-type perturbation space gradually becomes easier for the evolving \Repairer to solve.
This trend reveals both the effectiveness and the limitation of the current self-play curriculum: typed perturbations provide a useful learning signal, but a richer action vocabulary may be needed once the \Repairer becomes strong enough to saturate the existing perturbation space. 
Finally, the validation reward improves consistently, suggesting that the verifier-guided rewards provide a stable learning signal.

\begin{table}[t]
\centering
\caption{\textbf{Transfer across model scale and family.} The same perturbation and reward formulation improves both evaluation averages.}
\label{tab:backbones}
\small
\begin{tabular}{@{}lcccc@{}}
\toprule
& \multicolumn{2}{c}{Comp. Avg.} & \multicolumn{2}{c}{Broader Avg.} \\
\cmidrule(lr){2-3}\cmidrule(l){4-5}
Backbone & Base & \method & Base & \method \\
\midrule
Qwen3.5-2B & 83.22 & \textbf{86.85} & 58.77 & \textbf{61.03} \\
Qwen3.5-4B & 85.15 & \textbf{88.71} & 60.92 & \textbf{63.26} \\
InternVL3-2B & 82.19 & \textbf{85.34} & 58.16 & \textbf{60.99} \\
\bottomrule
\end{tabular}
\end{table}

\begin{figure}[t]
  \centering
  \includegraphics[width=\linewidth]{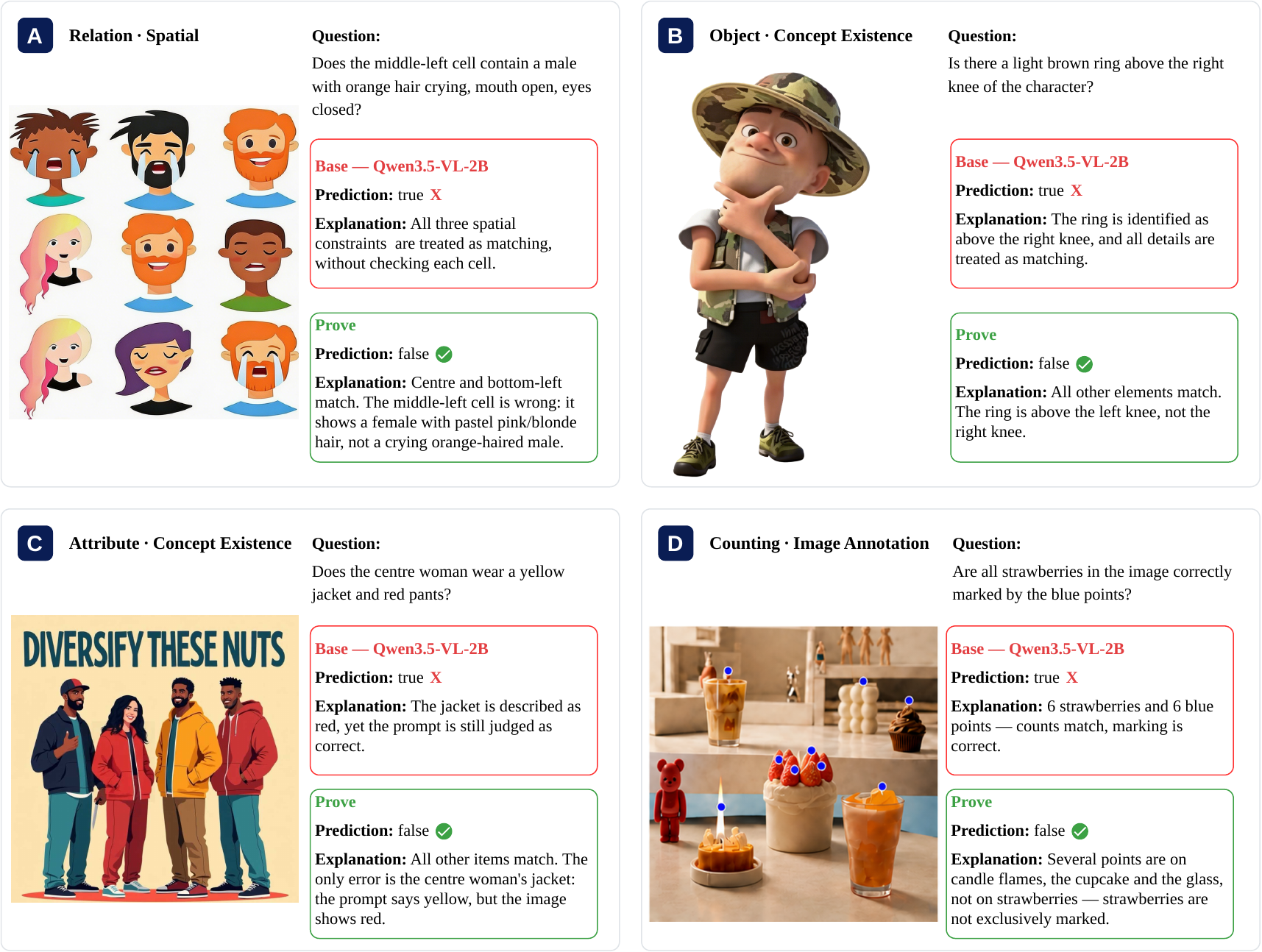}
  \caption{
  \textbf{Qualitative verifier examples.}
  Each panel compares the base Qwen3.5-2B model with \method on a fine-grained verification case.
  The examples cover relation, object existence, attribute, and counting errors.
  \method better localizes the relevant visual evidence and avoids accepting captions or answers that miss the targeted mismatch.
  }
  \label{fig:qualitative}
\end{figure}

\noindent\textbf{Axis-level gains.}
Figure~\ref{fig:axis_gains}(b) summarizes where the gains of \method come from by aggregating benchmark splits according to the visual-semantic capability they test.
The largest improvements appear on attribute and relation verification, which are directly aligned with the typed perturbations used by the \Mutator.
The full analysis is provided in Appendix~\ref{app:axis_gains}.

\noindent\textbf{Controlled loop ablations.}
Removing the \Repairer and second verification lowers the two averages by $2.61/1.72$ points; removing verifier feedback costs $1.68/1.19$ (Table~\ref{tab:controlled}).
Freezing the internal Verifier costs $1.73/1.22$, supporting verifier co-evolution.
Reward and role-sharing ablations also favor the full loop, while replacing DrGRPO with GRPO costs only $0.43/0.32$ (Appendix~\ref{app:controls}).

\noindent\textbf{Independent verification.}
Table~\ref{tab:full_audit} shows a $7.8$-point verifier accuracy gain and a $23.4$-point increase in fully valid trajectories, requiring a real mismatch, correct diagnosis, faithful minimal repair, and correct final judgment.
Fixed external benchmark averages also improve by $3.63/2.26$ points across training; their correlations with self-play pass rate are $0.9254/0.9022$ (Appendix~\ref{app:external_trajectory}).
These independent measurements support improved image-grounded correctness beyond internal role agreement.

\noindent\textbf{Generality and capability boundary.}
Table~\ref{tab:backbones} shows gains of $3.56/2.34$ on Qwen3.5-4B and $3.15/2.83$ on InternVL3-2B.
Three Qwen3.5-2B runs yield a compositional average of $86.73\pm0.27$, but ViVerBench, MMMU, and MathVista regress while MMBench improves and POPE remains stable (Appendix~\ref{app:seeds}).
Multi-edit curricula, unseen perturbation types, and caption filtering are studied in Appendices~\ref{app:generalization} and~\ref{app:caption_quality}. Figure~\ref{fig:qualitative} illustrates the learned verifier behavior. Additional failure analysis and task extensions are provided in Appendix~\ref{app:qualitative}.

\noindent\textbf{Limitations.}
\label{sec:limitations}
\method uses four types of localized caption perturbations, which align with the evaluated compositional axes and may saturate as the \Repairer improves.
Although independent audits support improved visual correctness, an evolving shared Verifier can still reinforce its own perceptual mistakes.
Training depends on faithful starting captions from one corpus and requires four role-conditioned calls per trajectory, increasing compute relative to static supervision.
Broader capability changes are benchmark-dependent, including regressions on ViVerBench, MMMU, and MathVista.
Task-specific self-play for VQA, grounding, and referring expressions remains untested.
Imperfect verifiers can also propagate errors when used for automated evaluation or filtering; these uses require independent validation.



\section{Conclusion}

We introduced \method, a self-play framework for training vision-language verifiers through a closed perturb--verify--repair--verify loop. 
Instead of relying on stronger external judges, fixed hard-negative datasets, or static preference annotations, \method converts ordinary clean image-caption pairs into adaptive verification episodes. 
A single shared MLLM acts as a \Mutator, Verifier, and \Repairer under different role prompts: the \Mutator creates localized image-text mismatches, the Verifier diagnoses them, and the \Repairer attempts to recover faithful captions from visual evidence and verifier feedback. 
This design turns verification into an operational self-improvement problem, where the model learns not only to judge image-text faithfulness but also to expose and correct its own fine-grained failure modes.
The resulting training signal is grounded, adaptive, and verifier-guided. 
Typed perturbations over objects, attributes, counts, and relations provide controllable supervision for compositional understanding, while the two verification rounds produce rewards that update the same backbone used at test time. 
Experiments on Qwen3.5-2B, Qwen3.5-4B, and InternVL3-2B backbones show that \method improves fine-grained compositional understanding across multiple benchmarks. On the reference backbone, it also strengthens visual-detail sensitivity, with limited, benchmark-dependent changes on broader multimodal tasks.
These results suggest that self-play over perturbation, verification, and repair offers a practical alternative to static judge distillation for building stronger vision-language verifiers.

\section*{Acknowledgments}
This work was supported by the European Union’s Horizon Europe research and innovation programme under
grant agreement number 101214398 (ELLIOT).
\clearpage
\bibliographystyle{plainnat}
\bibliography{refs}

\clearpage
\appendix

\section{Additional Experimental Details}
\label{app:details}

\subsection{Training Hyperparameters}
\label{app:hparams}

Table~\ref{tab:hparams} summarizes the main hyperparameters used for the Qwen3.5-2B reference run. 
The same shared backbone is used for the \Mutator, Verifier, and \Repairer roles, with role identity specified by prompt templates.

\begin{table}[ht]
\caption{
Training hyperparameters for the Qwen3.5-2B reference run.
}
\label{tab:hparams}
\centering
\small
\begin{tabular}{ll}
\toprule
Hyperparameter & Value \\
\midrule
Backbone & Qwen3.5-2B \\
Training data & 48k image-caption pairs \\
Validation data & 5k image-caption pairs \\
max prompt length / response length & 4096 / 1024 \\
rollout batch size & 48 \\
rollout size $n$ & 3 \\
sampling temperature / top-p / top-k & 0.7 / 0.8 / 20 \\
presence penalty & 1.5 \\
actor learning rate & $5\times10^{-7}$ \\
global batch size & 48 \\
PPO clip range & $[0.20, 0.22]$ \\
KL coefficient $\beta$ & $0.005$ \\
invalid-action penalty & $-0.3$ \\
precision & BF16 \\
hardware & 8 NVIDIA A100 GPUs \\
reported checkpoint & best validation checkpoint \\
\bottomrule
\end{tabular}
\end{table}

\subsection{Additional Benchmark Results}
\label{app:additional_results}

Table~\ref{tab:extended_checkpoints} reports checkpoint-level results on additional benchmarks~\citep{zhang2025omniverifier, liu2024ocrbench, yue2024mmmu, lu2024mathvista, kembhavi2016ai2d}. 
The checkpoints are denoted as \method@450, \method@600, and \method@750 according to the training step. 
The main paper reports the best validation checkpoint, while this table provides the full trajectory for completeness.

\begin{table}[ht]
\caption{
Extended checkpoint results for Qwen3.5-2B.
\method@450, \method@600, and \method@750 denote checkpoints at steps 450, 600, and 750, respectively.
}
\label{tab:extended_checkpoints}
\centering
\small
\begin{tabular}{lcccc}
\toprule
Benchmark & Base & \method@450 & \method@600 & \method@750 \\
\midrule
Winoground (2-AFC) & 69.50 & 74.50 & 74.75 & \textbf{75.62} \\
ARO macro & 73.57 & 78.01 & 78.27 & \textbf{78.66} \\
SugarCrepe micro & 93.76 & 94.24 & \textbf{94.28} & 94.17 \\
VL-CheckList 2-axis avg & 79.80 & 83.58 & 83.87 & \textbf{84.20} \\
POPE F1 & 89.57 & 89.52 & 89.58 & \textbf{89.66} \\
MMVP Pair Acc. & 56.00 & 59.33 & 57.33 & \textbf{60.00} \\
ViVerBench (rule) & \textbf{55.16} & 52.81 & 52.79 & 52.98 \\
MMBench-EN dev & 71.31 & 72.16 & 71.65 & \textbf{72.34} \\
OCRBench & 84.50 & -- & \textbf{84.90} & \textbf{84.90} \\
MMMU val & \textbf{40.78} & -- & 40.11 & 40.44 \\
MathVista testmini & 73.80 & \textbf{74.50} & 73.60 & 72.80 \\
AI2D & 75.26 & -- & 75.71 & \textbf{75.91} \\
\bottomrule
\end{tabular}
\end{table}

Table~\ref{tab:vlchecklist} reports the VL-CheckList axis-level results. 
The gains are consistent across attribute, object, and relation axes, showing that \method improves localized image-text verification rather than only a single benchmark score.

\begin{table}[ht]
\caption{
VL-CheckList axis-level results on Qwen3.5-2B.
}
\label{tab:vlchecklist}
\centering
\small
\begin{tabular}{lcccc}
\toprule
Metric & Base & \method@450 & \method@600 & \method@750 \\
\midrule
Attribute axis & 76.75 & 80.72 & 81.13 & \textbf{81.46} \\
Object axis & 91.20 & -- & -- & \textbf{94.14} \\
Relation axis & 82.85 & 86.45 & 86.60 & \textbf{86.95} \\
3-axis average & 83.60 & -- & -- & \textbf{87.52} \\
\bottomrule
\end{tabular}
\end{table}

\subsection{Reward Design Discussion}
\label{app:reward_design}

The verifier-guided rewards in \method are designed to couple two complementary goals: exposing verifier weaknesses and recovering faithful captions. 
For the \Mutator, the reward contains an exposure term, a hardness term, and an edit-cost penalty. 
The exposure term encourages perturbations that the current verifier fails to reject, while the hardness term encourages perturbations that remain challenging after repair. 
The cost term discourages overly wide edits. Actions rejected by the deterministic editor receive a fixed penalty instead of $C_M$ (Appendix~\ref{app:interfaces}).

For the \Repairer, the reward contains an acceptance term, a judgment-flip term, and a repair-cost penalty. 
The acceptance term encourages repaired captions that the verifier judges faithful, while the judgment-flip term rewards a change in the verifier decision, $\mathbb{1}[y_1\ne y_2]$; the separate acceptance term favors a faithful accepted repair.
The repair-cost penalty discourages unnecessary rewriting and encourages localized correction.

Table~\ref{tab:reward_design_compare} summarizes the role of each reward component.

\begin{table}[ht]
\caption{
Verifier-guided reward components in \method.
}
\label{tab:reward_design_compare}
\centering
\small
\begin{tabular}{lll}
\toprule
Role & Component & Purpose \\
\midrule
\Mutator & $R_{\mathrm{expose}}(y_1)$ & Expose verifier blind spots \\
\Mutator & $R_{\mathrm{hard}}(y_2)$ & Prefer perturbations that remain challenging after repair \\
\Mutator & $C_M(a_M)$ & Penalize overly wide edits \\
\Mutator & invalid-action penalty & Fixed penalty when the editor rejects the action \\
\Repairer & $R_{\mathrm{accept}}(y_2)$ & Reward repaired captions accepted by the verifier \\
\Repairer & $R_{\mathrm{flip}}(y_1,y_2)$ & Reward a change in the verifier judgment \\
\Repairer & $C_R(p',\tilde p)$ & Penalize unnecessary rewriting \\
\bottomrule
\end{tabular}
\end{table}

\subsection{Compute Cost}
\label{app:compute}

The Qwen3.5-2B reference run is trained on a single node with 8 NVIDIA A100 GPUs using BF16 mixed precision. 
Each self-play episode executes the perturb--verify--repair--verify trajectory with one shared backbone under different role prompts. 
The main run uses rollout size $n=3$ and global batch size 48; the reported trajectory includes checkpoints through step 750.
The reference run takes 11 hours on 8 A100 GPUs, totaling 88 GPU-hours.
We will release the training and evaluation code, including launch scripts and prompt templates, to facilitate reproducibility.

We profile all complete pipelines on the same hardware and 48k source pairs (Table~\ref{tab:compute_comparison}).
For \method, supervision is generated online and is included in target-training time.
Its four role-conditioned calls per trajectory produce more generated tokens than either static negatives or offline external-judge distillation.
Thus, the benefit is adaptive supervision without a separate teacher or fixed negative set, rather than universally lower raw compute.

\begin{table}[htbp]
\centering
\caption{Complete pipeline compute costs. Generation and training costs are GPU-hours. Online generation for \method is included in its 88.0 GPU-hours.}
\label{tab:compute_comparison}
\small
\begin{tabular}{@{}lcccc@{}}
\toprule
Method & \makecell{Supervision\\generation} & \makecell{Target\\training} & \makecell{Total\\GPU-hours} & \makecell{Generated\\tokens/source} \\
\midrule
Fixed typed negatives & 4.8 & 43.2 & 48.0 & 46 \\
External-judge distillation & 27.6 & 45.3 & 72.9 & 214 \\
\method & Online & 88.0 & 88.0 & 1,126 \\
\bottomrule
\end{tabular}
\end{table}

\subsection{Illustrative Examples of the \method Pipeline}
\label{app:evo_examples}

To complement the method description in Section~\ref{sec:method}, we provide end-to-end examples for the four supported \Mutator perturbation types: \textbf{count}, \textbf{attribute}, \textbf{relation}, and \textbf{object}. 
Each example follows the perturb--verify--repair--verify trajectory:
\[
(x, p)
\xrightarrow{\text{\Mutator}}
p'
\xrightarrow{\text{Verifier}}
(y_1,r_1)
\xrightarrow{\text{\Repairer}}
\tilde p
\xrightarrow{\text{Verifier}}
(y_2,r_2).
\]

The \Mutator emits a localized edit of the form $(\texttt{edit\_type}, \texttt{target\_text}, \texttt{replacement})$, which is applied to the clean caption $p$ to obtain the perturbed caption $p'$. 
The Verifier first judges whether $p'$ matches the image and returns a binary decision $y_1$ with textual feedback $r_1$. 
Conditioned on the image, the perturbed caption, and this feedback, the \Repairer directly generates a repaired caption $\tilde p$. 
The Verifier is then invoked again to obtain $(y_2,r_2)$. 
These examples illustrate how \method converts clean image-caption pairs into grounded hard negatives and repair targets without requiring a stronger external judge.

\needspace{12\baselineskip}
\subsubsection{Count}

\begin{tcolorbox}[title = {Illustrative \method Pipeline -- Count Action}, breakable, fontupper=\small, fonttitle=\small]

\vspace{\medskipamount}
{\includegraphics[width=0.4\linewidth]{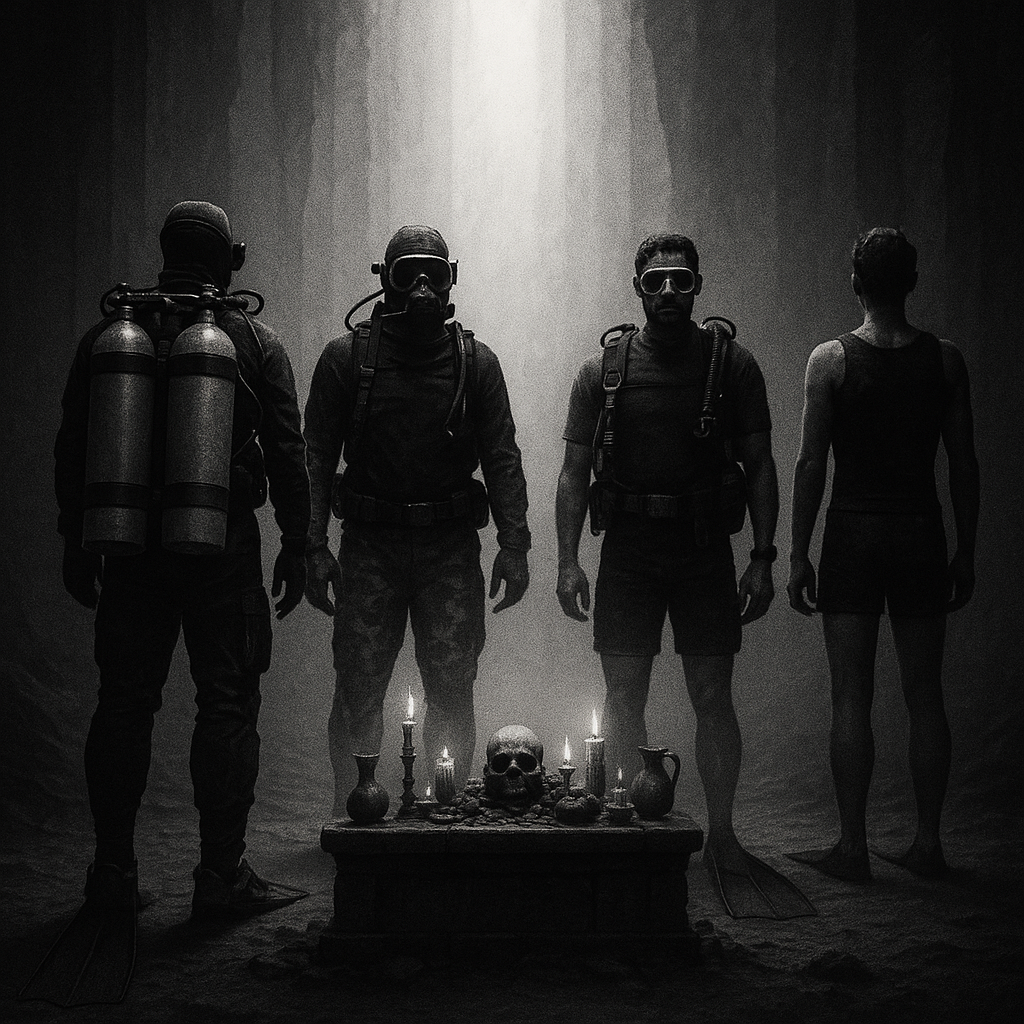}}
\vspace{\medskipamount}

\noindent\textbf{Original prompt $p$ (label $=1$):}\\
\textit{``A front view shot, utilizing strong vertical lines for composition, captures \textcolor{mydarkgreen}{four divers} before a central altar laden with mysterious offerings, their varied forms: the first, a technical specialist with a backplate and wing, twin steel tanks, a sealed drysuit, separate thick neoprene hood, and powerful blade fins; the second in a full wetsuit employing a sidemount BCD with a single tank, their face obscured by a full-face mask with integrated regulator, and distinguished by camouflage patterned fins; a third clearly recreational, with a jacket-style BCD, single aluminum tank, low-volume dual-lens mask, and standard paddle fins; and the fourth minimally equipped in a shorty wetsuit and full-foot fins---dramatically silhouetted by a strong, warm backlight yet softly defined by diffused ambient light, all rendered in an often high-contrast black and white artistic style.''}

\noindent\rule{\linewidth}{0.5pt}

\noindent\textbf{\Mutator action:}
{\scriptsize\ttfamily \{"edit\_type": "count", "target\_text": "four divers", "replacement": "five divers"\}}

\noindent\textbf{Perturbed prompt $p'$ (label $=0$):}\\
\textit{``A front view shot $\dots$ captures \textcolor{red}{five divers} before a central altar laden with mysterious offerings, their varied forms: the first $\dots$ and the fourth minimally equipped in a shorty wetsuit and full-foot fins---$\dots$''}

\noindent\textbf{Verifier (1st pass) on $(x, p')$:}
{\scriptsize\ttfamily
\textbf{\textcolor{blue}{<think>}} The user wants me to verify whether the description matches the image. \textbf{Analyze the image:} four divers stand in front of a central stone altar; the altar holds a skull, several candles, and small jugs; a strong vertical backlight produces silhouettes against a high-contrast black-and-white scene$\dots$ \textbf{Analyze the description:} ``$\dots$ captures \emph{five} divers $\dots$'' --- \textbf{Mismatch}: the image clearly shows \textbf{four} divers, not five. The altar with offerings, the warm backlight, and the B\&W styling all align with the rest of the prompt; only the count claim is violated$\dots$ \textbf{Conclusion:} the description does not accurately match the image. \textbf{\textcolor{blue}{</think>}}\\
\{ "answer": false,  "explanation": "The image contains four divers around the altar, not five." \}
}

\vspace{2pt}
$\Rightarrow y_1 = \texttt{false},\ r_1 = $ ``The image contains four divers around the altar, not five.''

\noindent\rule{\linewidth}{0.5pt}

\noindent\textbf{\Repairer output $\tilde{p}$:}\\
\textit{``A front view shot, utilizing strong vertical lines for composition, captures \textcolor{mydarkgreen}{four divers} before a central altar laden with mysterious offerings $\dots$''} \quad (recovers the count slot in $p$)

\noindent\textbf{Verifier (2nd pass) on $(x, \tilde{p})$:}
{\scriptsize\ttfamily
\textbf{\textcolor{blue}{<think>}} \textbf{Re-analyze the image:} four figures stand in a row before the central altar with a skull, candles, and small jugs; the vertical god-ray and high-contrast B\&W styling are unchanged from the previous pass$\dots$ \textbf{Re-analyze the description:} ``$\dots$ captures \emph{four} divers $\dots$'' --- the count is now correct, and the altar, the warm backlight, and the B\&W rendering all line up with the description$\dots$ \textbf{Conclusion:} modulo a few gear-level paraphrases the description is consistent with the image. \textbf{\textcolor{blue}{</think>}}\\
\{ "answer": true, "explanation": "Four divers are visible around the altar, matching the description." \}
}

\vspace{2pt}
$\Rightarrow y_2 = \texttt{true}$. Alignment score $s \approx 0.995$ ($\tilde p$ near-identical to $p$, count slot fully restored).

\end{tcolorbox}

\needspace{12\baselineskip}
\subsubsection{Attribute}

\begin{tcolorbox}[title = {Illustrative \method Pipeline -- Attribute Action}, breakable, fontupper=\small, fonttitle=\small]

\vspace{\medskipamount}
{\includegraphics[width=0.4\linewidth]{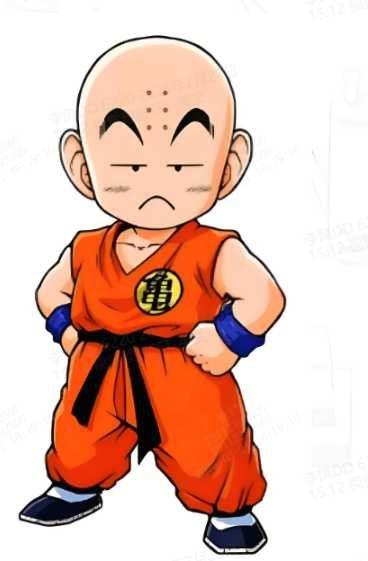}}
\vspace{\medskipamount}

\noindent\textbf{Original prompt $p$ (label $=1$):}\\
\textit{``A bald chibi character stands with hands on his hips, wearing an \textcolor{mydarkgreen}{orange martial arts uniform} with a yellow circular logo on the chest, a black belt around the waist, and blue wristbands on both arms.''}

\noindent\rule{\linewidth}{0.5pt}

\noindent\textbf{\Mutator action:}
{\scriptsize\ttfamily \{"edit\_type": "attribute", "target\_text": "an orange martial arts uniform", "replacement": "a green martial arts uniform"\}}

\noindent\textbf{Perturbed prompt $p'$ (label $=0$):}\\
\textit{``A bald chibi character stands with hands on his hips, wearing a \textcolor{red}{green martial arts uniform} with a yellow circular logo on the chest, a black belt around the waist, and blue wristbands on both arms.''}

\noindent\textbf{Verifier (1st pass) on $(x, p')$:}
{\scriptsize\ttfamily
\textbf{\textcolor{blue}{<think>}} The user wants me to verify whether the description matches the image. \textbf{Analyze the image:} a bald chibi character stands with hands on his hips, wearing an unmistakably \emph{orange} martial-arts gi; the chest carries a yellow circular Dragon-Ball-style logo, a black belt is tied at the waist, and blue wristbands are visible on both wrists$\dots$ \textbf{Analyze the description:} ``$\dots$ wearing a \emph{green} martial arts uniform $\dots$ yellow circular logo $\dots$ black belt $\dots$ blue wristbands $\dots$'' --- \textbf{Mismatch} on the uniform colour (\emph{green} vs.\ \emph{orange}); the logo, belt, and wristbands all match$\dots$ \textbf{Conclusion:} the description has a single colour error on the uniform. \textbf{\textcolor{blue}{</think>}}\\
\{ "answer": false, "explanation": "The uniform is orange, not green." \}
}

\vspace{2pt}
$\Rightarrow y_1 = \texttt{false},\ r_1 = $ ``The uniform is orange, not green.''

\noindent\rule{\linewidth}{0.5pt}

\noindent\textbf{\Repairer output $\tilde{p}$:}\\
\textit{``A bald chibi character stands with hands on his hips, wearing an \textcolor{mydarkgreen}{orange martial arts uniform} with a yellow circular logo on the chest, a black belt around the waist, and blue wristbands on both arms.''} \quad (identical to $p$, attribute slot recovered)

\noindent\textbf{Verifier (2nd pass) on $(x, \tilde{p})$:}
{\scriptsize\ttfamily
\textbf{\textcolor{blue}{<think>}} \textbf{Re-analyze the image:} bald chibi character, hands on hips, orange gi, yellow circular logo on the chest, black belt at the waist, blue wristbands on both wrists$\dots$ \textbf{Re-analyze the description:} ``$\dots$ orange martial arts uniform $\dots$ yellow circular logo $\dots$ black belt $\dots$ blue wristbands $\dots$'' --- every attribute slot now agrees with the image, and no further mismatch remains$\dots$ \textbf{Conclusion:} the description is an accurate match for the image. \textbf{\textcolor{blue}{</think>}}\\
\{ "answer": true, "explanation": "All visual attributes (orange gi, yellow logo, black belt, blue wristbands) match the description." \}
}

\vspace{2pt}
$\Rightarrow y_2 = \texttt{true}$. Alignment score $s \approx 0.99$ (single attribute slot restored).

\end{tcolorbox}

\needspace{12\baselineskip}
\subsubsection{Relation}

\begin{tcolorbox}[title = {Illustrative \method Pipeline -- Relation Action}, breakable, fontupper=\small, fonttitle=\small]

\vspace{\medskipamount}
{\includegraphics[width=0.45\linewidth]{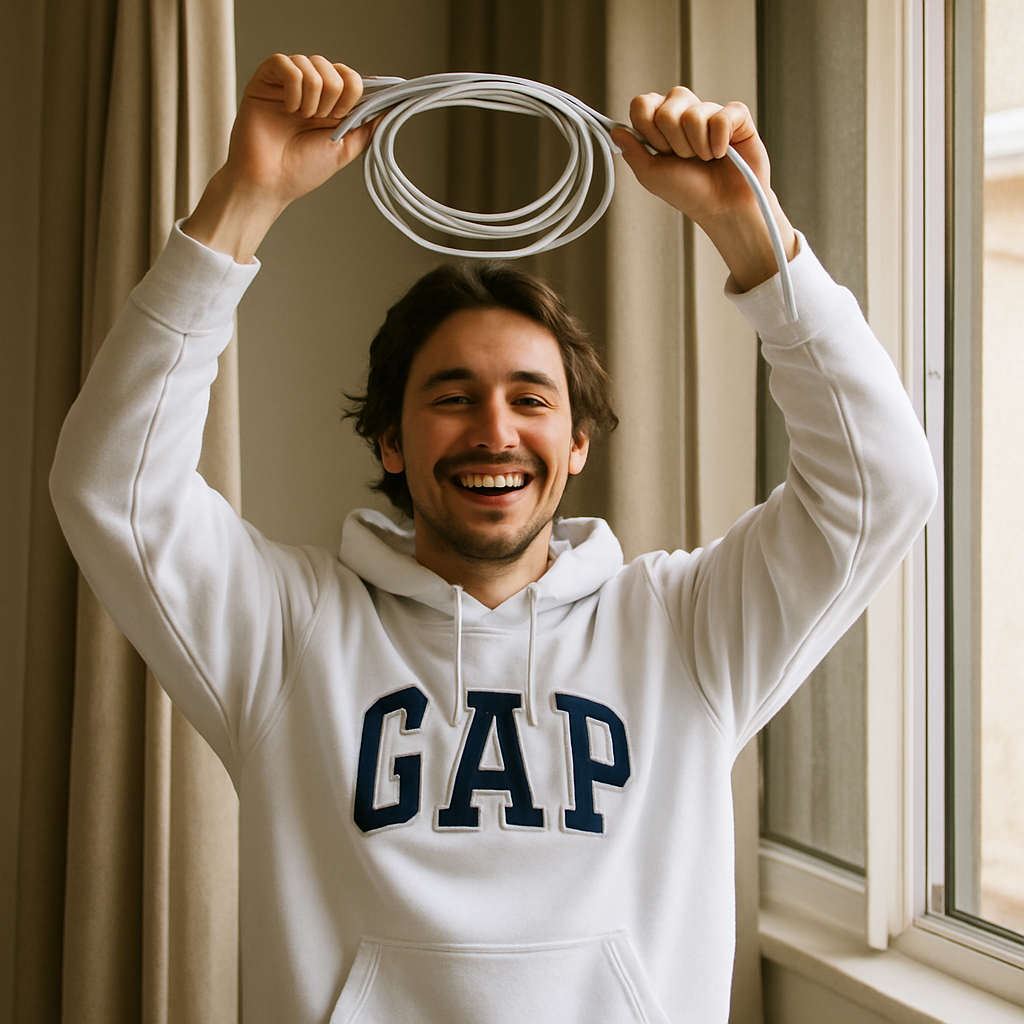}}
\vspace{\medskipamount}

\noindent\textbf{Original prompt $p$ (label $=1$):}\\
\textit{``A person wearing a white GAP hoodie is \textcolor{mydarkgreen}{standing indoors} with arms raised, holding a coiled white cable in both hands, near an open window with curtains.''}

\noindent\rule{\linewidth}{0.5pt}

\noindent\textbf{\Mutator action:}
{\scriptsize\ttfamily \{"edit\_type": "relation", "target\_text": "standing indoors", "replacement": "standing outside"\}}

\noindent\textbf{Perturbed prompt $p'$ (label $=0$):}\\
\textit{``A person wearing a white GAP hoodie is \textcolor{red}{standing outside} with arms raised, holding a coiled white cable in both hands, near an open window with curtains.''}

\noindent\textbf{Verifier (1st pass) on $(x, p')$:}
{\scriptsize\ttfamily
\textbf{\textcolor{blue}{<think>}} The user wants me to verify whether the description matches the image. \textbf{Analyze the image:} a young man with dark hair and a beard wears a white \texttt{GAP} hoodie; his arms are raised above his head and his hands hold a coiled white cable; behind him, heavy beige curtains hang on the left and a white-framed window is visible on the right, with soft diffuse interior lighting$\dots$ \textbf{Analyze the description:} ``$\dots$ \emph{standing outside} $\dots$'' --- \textbf{Mismatch}: the curtains, the window frame, and the indoor lighting all indicate an interior space, not an outdoor one. The hoodie, raised arms, and coiled cable do match$\dots$ \textbf{Conclusion:} the spatial relation in the prompt is inverted with respect to the image. \textbf{\textcolor{blue}{</think>}}\\
\{ "answer": false, "explanation": "The person is standing indoors, not outside." \}
}

\vspace{2pt}
$\Rightarrow y_1 = \texttt{false},\ r_1 = $ ``The person is standing indoors, not outside.''

\noindent\rule{\linewidth}{0.5pt}

\noindent\textbf{\Repairer output $\tilde{p}$:}\\
\textit{``A person wearing a white GAP hoodie is \textcolor{mydarkgreen}{indoors} with arms raised, holding a coiled white cable in both hands, near an open window with curtains.''}

\noindent\textbf{Verifier (2nd pass) on $(x, \tilde{p})$:}
{\scriptsize\ttfamily
\textbf{\textcolor{blue}{<think>}} \textbf{Re-analyze the image:} indoor scene with beige curtains on the left and a white-framed window on the right; a man in a white \texttt{GAP} hoodie raises his arms above his head and holds a coiled white cable in both hands$\dots$ \textbf{Re-analyze the description:} ``$\dots$ \emph{indoors} with arms raised, holding a coiled white cable $\dots$ near an open window with curtains'' --- every clause aligns with the image: the indoor relation, the raised arms, the coiled cable, and the window with curtains$\dots$ \textbf{Conclusion:} the description accurately matches the image. \textbf{\textcolor{blue}{</think>}}\\
\{ "answer": true, "explanation": "The person is indoors wearing a white GAP hoodie with arms raised holding a cable, near an open window with curtains." \}
}

\vspace{2pt}
$\Rightarrow y_2 = \texttt{true}$. Alignment score $s = 0.958$ (relation slot recovered; ``standing'' dropped, otherwise close to $p$).

\end{tcolorbox}

\needspace{12\baselineskip}
\subsubsection{Object}

\begin{tcolorbox}[title = {Illustrative \method Pipeline -- Object Action}, breakable, fontupper=\small, fonttitle=\small]

\vspace{\medskipamount}
{\includegraphics[width=0.35\linewidth]{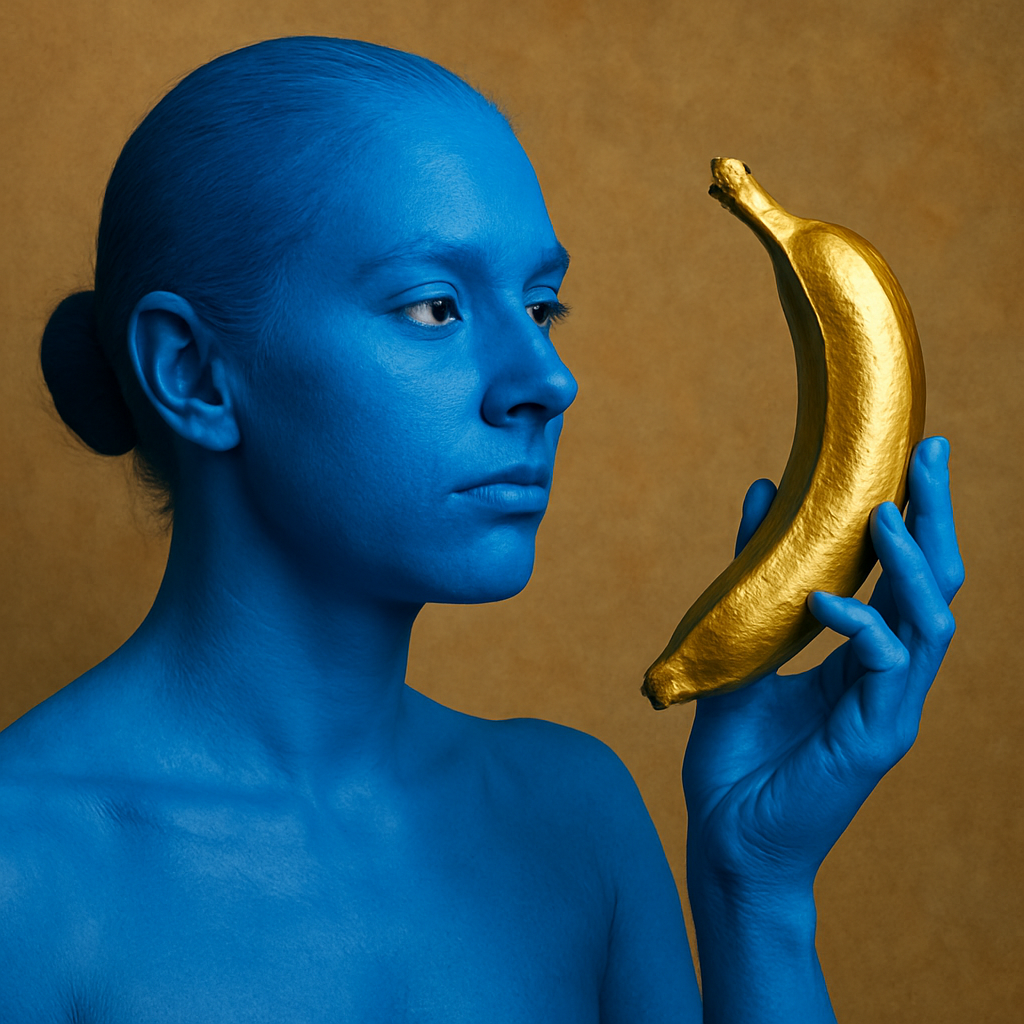}}
\vspace{\medskipamount}

\noindent\textbf{Original prompt $p$ (label $=1$):}\\
\textit{``A blue-painted woman holds a single \textcolor{mydarkgreen}{gold banana} up to her face against a warm tan backdrop, in a stark studio portrait style.''}

\noindent\rule{\linewidth}{0.5pt}

\noindent\textbf{\Mutator action:}
{\scriptsize\ttfamily \{"edit\_type": "object", "target\_text": "gold banana", "replacement": "gold pear"\}}

\noindent\textbf{Perturbed prompt $p'$ (label $=0$):}\\
\textit{``A blue-painted woman holds a single \textcolor{red}{gold pear} up to her face against a warm tan backdrop, in a stark studio portrait style.''}

\noindent\textbf{Verifier (1st pass) on $(x, p')$:}
{\scriptsize\ttfamily
\textbf{\textcolor{blue}{<think>}} The user wants me to verify whether the description matches the image. \textbf{Analyze the image:} a woman with blue-painted skin holds a gold-coloured object up to her face against a warm tan studio backdrop; the object's silhouette is long, curved, and tapered at one end---unambiguously a banana, not the round, broader-based shape of a pear$\dots$ \textbf{Analyze the description:} ``$\dots$ holds a single \emph{gold pear} up to her face $\dots$'' --- \textbf{Mismatch} on the fruit identity. The blue body paint, the warm tan backdrop, and the stark studio lighting all align with the prompt$\dots$ \textbf{Conclusion:} the object slot is incorrect (\emph{pear} where the image shows a \emph{banana}). \textbf{\textcolor{blue}{</think>}}\\
\{ "answer": false, "explanation": "The object held is a gold banana, not a gold pear." \}
}

\vspace{2pt}
$\Rightarrow y_1 = \texttt{false},\ r_1 = $ ``The object held is a gold banana, not a gold pear.''

\noindent\rule{\linewidth}{0.5pt}

\noindent\textbf{\Repairer output $\tilde{p}$:}\\
\textit{``A blue-painted woman holds a single \textcolor{mydarkgreen}{gold banana} up to her face $\dots$''} \quad (identical to $p$, object slot recovered)

\noindent\textbf{Verifier (2nd pass) on $(x, \tilde{p})$:}
{\scriptsize\ttfamily
\textbf{\textcolor{blue}{<think>}} \textbf{Re-analyze the image:} a blue-painted woman holds a single gold banana up near her face; the background is a textured warm tan/brown, and the lighting and framing are clearly studio-portrait style$\dots$ \textbf{Re-analyze the description:} ``$\dots$ \emph{gold banana} $\dots$ warm tan backdrop $\dots$ stark studio portrait style'' --- the fruit identity, the colour, the backdrop, and the studio styling all match$\dots$ \textbf{Conclusion:} the description accurately reflects the visual elements of the image. \textbf{\textcolor{blue}{</think>}}\\
\{ "answer": true, "explanation": "The blue-painted woman is holding a gold banana near her face against a warm tan backdrop, matching the prompt." \}
}

\vspace{2pt}
$\Rightarrow y_2 = \texttt{true}$. Alignment score $s \approx 0.99$ (object slot fully restored).

\end{tcolorbox}

\paragraph{Takeaways.} Across the four edit types, the same minimal-edit principle holds: the \Mutator flips exactly one semantic slot (\texttt{count}, \texttt{attribute}, \texttt{relation}, or \texttt{object}) via a structured edit, the first Verifier pass localises the mismatch via a short rationale, and the \Repairer uses that rationale to directly generate a faithful caption $\tilde p$ that, modulo at most a few-token paraphrase (e.g.\ ``standing indoors''~$\to$~``indoors''), matches the original $p$. The second Verifier pass closes the loop and produces the success signal $y_2$ that drives the \Repairer reward $r_R$ (Eq.~\ref{eq:repairer_reward_inst}). Because perturbation is a discrete typed action while repair is free-form caption generation, the two roles are not symmetric: the \Mutator is constrained to expose verifier blind spots along a known axis, whereas the \Repairer must recover the full caption from visual evidence and verifier feedback, which is what makes repair an operational form of verification.

\section{Controlled Supervision and Loop Ablations}
\label{app:controls}

\noindent\textbf{Matched supervision settings.}
All comparisons use the same Qwen3.5-2B initialization, 48k clean image-caption pairs, target-model optimization budget, and evaluation protocol.
Fixed typed negatives are generated once and reused throughout training.
External-judge distillation uses judgments and textual feedback from a frozen larger VLM.
Verifier self-training uses self-generated labels without the \Repairer or second verification.
This baseline isolates self-generated verifier supervision under a matched image-text protocol; it is not a reproduction of reasoning-oriented self-play methods with different tasks and supervision.
Table~\ref{tab:controlled} reports the matched supervision results.
The same optimization budget does not imply equal GPU-hours, since the methods execute different generation pipelines (Table~\ref{tab:compute_comparison}).

\subsection{Component and Reward Ablations}
\label{app:component_ablations}

Table~\ref{tab:full_ablations} isolates the role of static versus adaptive perturbations, the repair loop, verifier feedback, the optimizer, and reward components.
Removing the \Repairer and second verification produces the largest drop, $2.61/1.72$ points on the compositional/broader averages.
Removing feedback costs $1.68/1.19$ points, indicating that repair benefits from the verifier's diagnosis.
Both role rewards contribute: removing the \Mutator or \Repairer reward costs $1.82/1.27$ or $1.31/0.94$ points, respectively.
Removing the edit and repair costs reduces the averages by $0.77/0.60$ points.
Replacing DrGRPO with GRPO produces a smaller $0.43/0.32$ drop; the main gain therefore depends on the closed-loop supervision rather than the optimizer alone.

\begin{table}[htbp]
\centering
\caption{Complete component, optimizer, and reward ablations. All variants use the same source data, backbone initialization, target-model optimization budget, and evaluation protocol; the GRPO row changes only the optimization variant.}
\label{tab:full_ablations}
\small
\begin{tabular}{@{}lcc@{}}
\toprule
Training variant & Comp. Avg. & Broader Avg. \\
\midrule
Qwen3.5-2B & 83.22 & 58.77 \\
Fixed typed negatives & 85.02 & 59.68 \\
No \Repairer or second verification & 84.24 & 59.31 \\
\Repairer without verifier feedback & 85.17 & 59.84 \\
GRPO instead of DrGRPO & 86.42 & 60.71 \\
Without \Mutator reward & 85.03 & 59.76 \\
Without \Repairer reward & 85.54 & 60.09 \\
Without edit/repair costs & 86.08 & 60.43 \\
\rowcolor{blue!4}
\method & \textbf{86.85} & \textbf{61.03} \\
\bottomrule
\end{tabular}
\end{table}

\subsection{Role Configuration and Verifier Updating}
\label{app:role_ablations}

Table~\ref{tab:role_configurations} compares shared role prompts with role-specific adapters and frozen Verifier variants.
Freezing the internal Verifier lowers the compositional/broader averages by $1.73/1.22$ points, supporting verifier co-evolution.
Role-specific adapters remain competitive but are $0.44/0.25$ points below full parameter sharing.
A frozen external Verifier is stronger than the frozen internal Verifier, but remains $0.78/0.49$ points below the shared loop.
This frozen external Verifier operates online inside the self-play loop, whereas external-judge distillation uses offline teacher judgments as target supervision.
These are distinct comparisons.

\begin{table}[htbp]
\centering
\caption{Role configuration and verifier-update ablations.}
\label{tab:role_configurations}
\small
\begin{tabular}{@{}lcc@{}}
\toprule
Role configuration & Comp. Avg. & Broader Avg. \\
\midrule
Shared backbone, role prompts & \textbf{86.85} & \textbf{61.03} \\
Shared backbone, role-specific adapters & 86.41 & 60.78 \\
Frozen internal Verifier & 85.12 & 59.81 \\
Frozen external Verifier & 86.07 & 60.54 \\
\bottomrule
\end{tabular}
\end{table}

\section{Independent Verification and Repeated-Run Analysis}
\label{app:independent}

\subsection{External Trajectory Audit}
\label{app:external_audit}

We audit a held-out trajectory set using labels produced outside the self-play loop.
These labels are never used for training, reward computation, mutation generation, validation, or checkpoint selection.
The initial Qwen3.5-2B model and \method use the same role prompts and deterministic editor.
Table~\ref{tab:full_audit} reports verifier accuracy on a fixed set and the quality of the generated trajectories.

A valid localized perturbation introduces a real image-caption mismatch confined to the intended edit.
A correct verifier judgment identifies whether the caption is faithful to the image.
A faithful minimal repair corrects the mismatch without introducing unsupported changes.
A fully valid trajectory requires all four stages to be correct: a real localized mismatch, a correct diagnosis, a faithful minimal repair, and a correct final judgment.
The fully valid rate increases by $23.4$ points, while fixed-set verifier accuracy increases by $7.8$ points.
These external measurements support improved visual correctness, while the shared-model setting still leaves possible self-verification bias (Section~\ref{sec:limitations}).

\begin{table}[htbp]
\centering
\caption{Complete external audit. All entries are percentages. Fixed-set verifier accuracy and trajectory-quality rates are separate audit measurements.}
\label{tab:full_audit}
\small
\begin{tabular}{@{}lcc@{}}
\toprule
External audit & Base & \method \\
\midrule
Fixed-set verifier accuracy & 75.8 & \textbf{83.6} \\
Valid localized perturbations & 82.4 & \textbf{92.1} \\
Correct verifier judgments & 71.5 & \textbf{84.7} \\
Faithful minimal repairs & 55.2 & \textbf{73.4} \\
Fully valid trajectories & 40.8 & \textbf{64.2} \\
\bottomrule
\end{tabular}
\end{table}

\subsection{External Benchmark Trajectory}
\label{app:external_trajectory}

\begin{figure}[htbp]
\centering
\includegraphics[width=\linewidth]{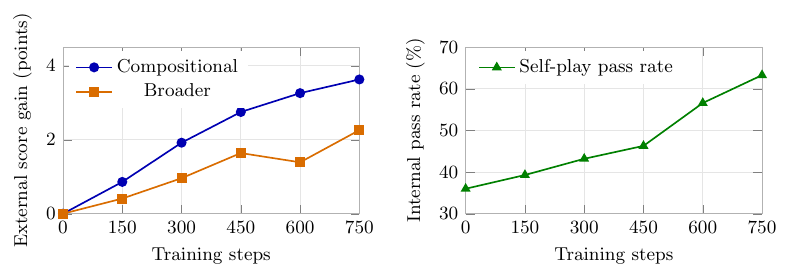}
\caption{\textbf{External benchmark performance and internal self-play pass rate.} Each marker is one reported checkpoint; connecting lines guide the eye. Gains in the left panel are measured against the base checkpoint. The right panel reports internal pass rate separately, preserving the distinction between independent accuracy and role agreement.}
\label{fig:external_trajectory}
\end{figure}

We evaluate checkpoints on fixed, externally labeled benchmarks that are excluded from training, self-play rewards, mutation generation, validation, and checkpoint selection.
Table~\ref{tab:external_checkpoints} separates the internal self-play pass rate from externally measured benchmark performance.
The compositional average increases from $83.22$ to $86.85$, while the broader average increases from $58.77$ to $61.03$.
The broader trajectory is not monotonic: its step-600 value is below step 450, illustrating why internal agreement is not a substitute for independent evaluation.
Across the six reported checkpoints, Pearson correlations between pass rate and the compositional/broader averages are $0.9254/0.9022$.
These descriptive correlations summarize co-movement during training; they do not establish a causal relationship or a significance test.
Figure~\ref{fig:external_trajectory} plots the six reported measurements without adding intermediate observations.

\begin{table}[htbp]
\centering
\caption{Internal self-play pass rate and external benchmark averages across training. The pass rate is in percent; benchmark averages use the main-paper definitions.}
\label{tab:external_checkpoints}
\small
\begin{tabular}{@{}lccc@{}}
\toprule
Checkpoint & Self-play pass rate & Comp. Avg. & Broader Avg. \\
\midrule
Base & 36.0 & 83.22 & 58.77 \\
Step 150 & 39.3 & 84.08 & 59.18 \\
Step 300 & 43.2 & 85.14 & 59.73 \\
Step 450 & 46.3 & 85.97 & 60.41 \\
Step 600 & 56.6 & 86.48 & 60.16 \\
Step 750 & 63.3 & 86.85 & 61.03 \\
\bottomrule
\end{tabular}
\end{table}

\subsection{Repeated-Run Variability and Broader Capability}
\label{app:seeds}

We repeat Qwen3.5-2B training with three independent seeds.
Table~\ref{tab:seed_statistics} reports the mean and standard deviation across the three trained models.
These statistics measure training-run variability, not example-level uncertainty, confidence intervals, or statistical significance.
The reference-run results in Tables~\ref{tab:cross_model} and~\ref{tab:additional_benchmarks} and the checkpoint table in Appendix~\ref{app:additional_results} remain single-run measurements.

The core compositional average improves by $3.51$ points across seeds.
Broader changes are limited and task-dependent: ViVerBench regresses by $2.04$ points, MMMU and MathVista decrease by $0.39$ and $0.86$, MMBench improves by $0.95$, and POPE remains nearly unchanged.
Thus, \method substantially improves compositional verification without implying uniform preservation or improvement of all broader capabilities.

\begin{table}[htbp]
\centering
\caption{Results across three independent Qwen3.5-2B training runs. Change is the mean \method result minus the base score.}
\label{tab:seed_statistics}
\small
\begin{tabular}{@{}lccc@{}}
\toprule
Benchmark & Base & \method (mean $\pm$ std) & Change \\
\midrule
ViVerBench & 55.16 & $53.12\pm0.58$ & $-2.04$ \\
MMMU & 40.78 & $40.39\pm0.47$ & $-0.39$ \\
MathVista & 73.80 & $72.94\pm0.52$ & $-0.86$ \\
MMBench & 71.31 & $72.26\pm0.31$ & $+0.95$ \\
POPE & 89.57 & $89.64\pm0.09$ & $+0.07$ \\
Core compositional average & 83.22 & $86.73\pm0.27$ & $+3.51$ \\
\bottomrule
\end{tabular}
\end{table}

\section{Generality and Perturbation Curricula}
\label{app:generalization}

\subsection{Backbone Scale and Family}
\label{app:backbones}

We apply the same perturbation and reward formulation to Qwen3.5-4B and the scale-matched InternVL3-2B family.
Table~\ref{tab:backbones} in the main paper reports the full base and \method averages for all three backbones.
The Qwen3.5-4B gains are $3.56/2.34$ compositional/broader points, while InternVL3-2B gains $3.15/2.83$.
These results show that the gains extend beyond the 2B Qwen reference setting, although all runs still rely on one source training corpus.

\subsection{Leave-One-Type-Out Transfer}
\label{app:leave_one_type_out}

For each leave-one-type-out experiment, we remove the held-out type from the \Mutator action space and resample the remaining types to preserve the same episode count and optimization budget.
Table~\ref{tab:leave_one_type_out} reports performance on the held-out axis.
The retained gain is $(s_{\mathrm{LOO}}-s_{\mathrm{base}})/(s_{\mathrm{full}}-s_{\mathrm{base}})$.
These axis-level scores follow the capability aggregation used in Figure~\ref{fig:axis_gains}(b), rather than the overall averages in the main tables.
Leave-one-out training retains $48.2\%$--$52.4\%$ of the full-method gain across the three reported axes.
This supports cross-type transfer while showing that direct exposure supplies additional axis-specific benefit.

\begin{table}[htbp]
\centering
\caption{Leave-one-type-out transfer on the three reported axes. The held-out axis is absent from the training action space.}
\label{tab:leave_one_type_out}
\small
\begin{tabular}{@{}lcccc@{}}
\toprule
Held-out type & Base & Leave-one-out \method & Full \method & Retained gain \\
\midrule
Object & 86.1 & 87.2 & 88.2 & 52.4\% \\
Attribute & 78.1 & 81.1 & 84.1 & 50.0\% \\
Relation & 76.3 & 79.0 & 81.9 & 48.2\% \\
\bottomrule
\end{tabular}
\end{table}

\subsection{Single- and Multi-Edit Curricula}
\label{app:curricula}

Table~\ref{tab:curricula} compares one-edit, mixed one/two-edit, and two-edit curricula.
The mixed curriculum raises the broader average by $0.15$ points but lowers compositional performance and repair success.
Using two edits throughout further reduces repair success to $48.6\%$.
These results support single localized edits as the default because they simplify diagnosis, minimal repair, and credit assignment.
A multi-edit extension may benefit from explicit partial-credit rewards, which are not evaluated in these comparisons.

\begin{table}[htbp]
\centering
\caption{Perturbation curriculum comparison. Repair success is a percentage.}
\label{tab:curricula}
\small
\begin{tabular}{@{}lccc@{}}
\toprule
Curriculum & Comp. Avg. & Broader Avg. & Repair success \\
\midrule
One edit & \textbf{86.85} & 61.03 & \textbf{62.7} \\
Mixture of one/two edits & 86.06 & \textbf{61.18} & 57.4 \\
Two edits & 85.72 & 60.31 & 48.6 \\
\bottomrule
\end{tabular}
\end{table}

\section{Positive-Caption Quality}
\label{app:caption_quality}

We independently audit a random subset of the training pool, distinguishing fully faithful captions, minor unsupported details or ambiguity, and material image-caption mismatches.
Table~\ref{tab:caption_quality} reports the three fractions, which sum to $100\%$.
Most captions are faithful, while $1.7\%$ contain material mismatches.
We then remove low-confidence positives and repeat training (Table~\ref{tab:filtered_positives}).
The two averages increase by $0.23/0.12$ points, suggesting that the observed verification gain does not depend on retaining noisy positives.
This does not remove the method's dependence on sufficiently faithful reference captions.

\begin{table}[htbp]
\centering
\caption{Independent audit of a random subset of the positive-caption pool.}
\label{tab:caption_quality}
\small
\begin{tabular}{@{}lc@{}}
\toprule
Caption quality & Fraction \\
\midrule
Fully faithful & 90.1\% \\
Minor unsupported detail or ambiguity & 8.2\% \\
Material image-caption mismatch & 1.7\% \\
\bottomrule
\end{tabular}
\end{table}

\begin{table}[htbp]
\centering
\caption{Training after filtering low-confidence positive captions.}
\label{tab:filtered_positives}
\small
\begin{tabular}{@{}lcc@{}}
\toprule
Training pool & Comp. Avg. & Broader Avg. \\
\midrule
Original 48k & 86.85 & 61.03 \\
Filtered positives & \textbf{87.08} & \textbf{61.15} \\
\bottomrule
\end{tabular}
\end{table}

\section{Role Interfaces and Deterministic Editing}
\label{app:interfaces}

\noindent\textbf{Role-conditioned interfaces.}
The \Mutator receives the image and a faithful caption and returns a single structured action with fields \texttt{edit\_type}, \texttt{target\_text}, and \texttt{replacement}.
The type is one of \texttt{object}, \texttt{attribute}, \texttt{count}, or \texttt{relation}.
The deterministic editor applies the local replacement to construct the caption supplied to the first Verifier.
The Verifier receives the image and candidate caption and produces a binary faithfulness decision with textual feedback; the illustrative outputs in Appendix~\ref{app:evo_examples} use the fields \texttt{answer} and \texttt{explanation}.
The \Repairer receives the image, perturbed caption, first decision, and explanation, and directly generates a faithful caption with a minimal correction.
The same Verifier interface is used for the second verification.
The complete templates used in training are listed below.

\noindent\textbf{Prompt templates.}
\texttt{<image>} marks the image input.
Braced fields are filled at run time: \texttt{\{p\}} is the clean caption $p$, \texttt{\{caption\}} is the candidate caption ($p'$ in the first verification and $\tilde p$ in the second), \texttt{\{p'\}} is the perturbed caption, and \texttt{\{y1\}} and \texttt{\{r1\}} are the first Verifier decision and explanation.

\begin{tcolorbox}[title={\Mutator prompt template}, breakable, fontupper=\scriptsize\ttfamily, fonttitle=\small]
<image>\{p\}\par\medskip
You are a Mutator agent. Look at the image and the caption above, then propose a single minimal edit that turns the caption into a slightly incorrect description of the image.\par\medskip
Output exactly one JSON object on a single line, with no extra text, no markdown fences, no comments, no thinking trace:\par
\{"edit\_type": "<object|attribute|count|relation>", "target\_text": "<exact substring from the caption>", "replacement": "<modified text, must differ from target\_text>"\}\par\medskip
Rules:\par
- edit\_type is one of object, attribute, count, relation.\par
- target\_text MUST be an exact substring that occurs once in the caption.\par
- replacement MUST differ from target\_text.\par
- Keep the edit minimal and locally semantic.\par
- Output ONLY the JSON object, nothing else.
\end{tcolorbox}

\begin{tcolorbox}[title={Verifier prompt template (both verifications)}, breakable, fontupper=\scriptsize\ttfamily, fonttitle=\small]
<image>\{caption\}\par\medskip
You are a Verifier agent. Determine whether the caption faithfully describes the image.\par\medskip
First write the reasoning inside <think> </think>. Then output exactly one JSON object:\par
\{"answer": <true|false>, "explanation": "<one short sentence>"\}\par\medskip
Rules:\par
- answer is true only when every object, attribute, count, and relation in the caption matches the image.\par
- answer is false when any object, attribute, count, or relation is wrong.\par
- explanation names the mismatch, or states that the caption matches. Example: "The image contains four divers around the altar, not five."\par
- Judge only what is visible. Do not rewrite the caption.
\end{tcolorbox}

\begin{tcolorbox}[title={\Repairer prompt template}, breakable, fontupper=\scriptsize\ttfamily, fonttitle=\small]
<image>\{p'\}\par\medskip
You are a Repairer agent. The caption above may not faithfully describe the image. The verifier decision is y1=\{y1\}, and its explanation is: \{r1\}\par\medskip
Recover a caption that faithfully describes the image. Make the smallest correction needed and preserve the rest of the caption.\par\medskip
Output only the repaired caption, with no JSON, no markdown fences, no comments, and no thinking trace.
\end{tcolorbox}

\noindent\textbf{Output parsing.}
The \Mutator output is parsed as $a_M=(\tau,t,\delta)$ from \texttt{edit\_type}, \texttt{target\_text}, and \texttt{replacement}.
The Verifier output follows the format of Appendix~\ref{app:evo_examples}: a reasoning trace inside \texttt{<think>} tags followed by the JSON object.
\texttt{answer}$=$\texttt{true} gives $y=1$, \texttt{answer}$=$\texttt{false} gives $y=0$, and \texttt{explanation} is the textual feedback $r$; the two verifications yield $(y_1,r_1)$ and $(y_2,r_2)$.
The \Repairer output is the repaired caption $\tilde p$ itself rather than an edit action.

\noindent\textbf{Deterministic editing constraints.}
A valid action has a supported edit type and nonempty target and replacement strings.
The target must identify one unambiguous span in the caption, and the replacement must change it.
Unsupported types, missing or ambiguous spans, unchanged replacements, and edits that cannot be applied deterministically are invalid.
For a unique target span, the editor preserves the caption prefix and suffix and replaces only that span.
Algorithm~\ref{alg:editor} specifies this local-edit interface.
An invalid action receives the invalid-action penalty $-0.3$ in Table~\ref{tab:hparams} as its \Mutator reward, and $C_M$ is not computed.
Invalid edits do not establish a valid negative example.

\begin{algorithm}[htbp]
\caption{Deterministic single-span editor specification}
\label{alg:editor}
\begin{algorithmic}[1]
\Require Caption $p$, structured action $a_M=(\tau,t,\delta)$
\If{$a_M$ cannot be parsed, $\tau\notin\mathcal{T}$, or $t,\delta$ are missing or empty}
    \State \Return invalid action
\EndIf
\If{$t=\delta$ or $t$ does not identify exactly one span in $p$}
    \State \Return invalid action
\EndIf
\State Write $p=u\,t\,v$ using the unique target occurrence
\State \Return $p'=u\,\delta\,v$
\end{algorithmic}
\end{algorithm}

\noindent\textbf{Reward weights.}
Equations~\ref{eq:mutator_reward_inst} and~\ref{eq:repairer_reward_inst} instantiate Equations~\ref{eq:mutator_reward_concept} and~\ref{eq:repairer_reward_concept} with $R_{\mathrm{expose}}(y_1)=\lambda_1\mathbb{1}[y_1=1]$, $R_{\mathrm{hard}}(y_2)=\lambda_2\mathbb{1}[y_2=0]$, and $\lambda_M=\lambda_3$ for the \Mutator, and $R_{\mathrm{accept}}(y_2)=\beta_1\mathbb{1}[y_2=1]$, $R_{\mathrm{flip}}(y_1,y_2)=\beta_2\mathbb{1}[y_1\ne y_2]$, and $\lambda_R=\beta_3$ for the \Repairer.
Table~\ref{tab:reward_weights} lists the weights used in all reported runs.

\begin{table}[htbp]
\centering
\caption{Reward weights in Equations~\ref{eq:mutator_reward_inst} and~\ref{eq:repairer_reward_inst}.}
\label{tab:reward_weights}
\small
\begin{tabular}{@{}lcl@{}}
\toprule
Weight & Value & Term \\
\midrule
$\lambda_1$ & 1.0 & Perturbation not rejected by the first verification \\
$\lambda_2$ & 0.5 & Repaired caption still judged unfaithful \\
$\lambda_3$ & 0.3 & \Mutator edit cost $C_M$ \\
$\beta_1$ & 0.8 & Repair accepted by the second verification \\
$\beta_2$ & 0.7 & Judgment flip between the two verifications \\
$\beta_3$ & 0.3 & \Repairer rewriting cost $C_R$ \\
\bottomrule
\end{tabular}
\end{table}

\noindent\textbf{Cost terms.}
$C_M$ penalizes overly wide edit spans, normalized by the character length of the clean caption:
\begin{equation}
\label{eq:mutator_cost}
C_M(a_M)=2\cdot\frac{\mathrm{len}(t)}{\mathrm{len}(p)},
\end{equation}
with $C_M=0$ when $t$ or $p$ is empty.
$C_R$ penalizes unnecessary rewriting.
Let $T(\cdot)$ split a caption on whitespace and lowercase each token, and let $\mathrm{lev}$ be the token-level Levenshtein distance:
\begin{equation}
\label{eq:repairer_cost}
C_R(p',\tilde p)=\frac{\mathrm{lev}\big(T(p'),T(\tilde p)\big)}{\max\big(|T(p')|,|T(\tilde p)|\big)}.
\end{equation}
$C_R$ is $0$ for identical captions, approaches $1$ for a complete rewrite, and is set to $1$ when either caption has no tokens.

\noindent\textbf{Reward and diagnostic interpretation.}
Equations~\ref{eq:mutator_reward_inst} and~\ref{eq:repairer_reward_inst} specify the exposure, hardness, acceptance, judgment-flip, and cost terms.
The flip indicator $\mathbb{1}[y_1\ne y_2]$ rewards a decision change; it is not independently a guarantee of a visually correct repair.
In particular, an evolving Verifier can accept an incorrect caption, motivating the external audit in Appendix~\ref{app:external_audit}.
Repair success, structural alignment, and internal pass rate are training diagnostics, whereas the external benchmark and audit labels measure correctness independently.
For a group of $n=3$ binary outcomes, the empirical success fraction can take values $0,1/3,2/3,1$.
Its Bernoulli variance $\bar y(1-\bar y)$ is maximized at both $1/3$ and $2/3$; Figure~\ref{fig:dynamics} plots the lower finite-group reference.
This differs from the population Bernoulli variance maximum at $1/2$ and is a diagnostic reference, not an additional measurement or an external accuracy target.

\section{Axis-level Gains, Qualitative Behavior, and Task Extensions}
\label{app:qualitative}
\subsection{Axis-level Gains}
\label{app:axis_gains}
\noindent\textbf{Axis-level gains.}
Figure~\ref{fig:axis_gains}(b) summarizes where the gains of \method come from by aggregating benchmark splits according to the visual-semantic capability they test.
The largest improvements appear on attribute and relation verification, which are directly aligned with the typed perturbations used by the \Mutator.
This pattern supports the intended training mechanism: localized perturbations create visually checkable mismatches, forcing the Verifier to diagnose fine-grained inconsistencies and the \Repairer to recover faithful captions.
The positive gain on visual-detail sensitivity further suggests that verifier-guided repair improves localized evidence grounding beyond the core compositional benchmarks.
In contrast, object and color/order gains are smaller but still positive, partly because the base model already performs strongly on these axes.

\subsection{Qualitative Behavior and Failure Cases}
\noindent\textbf{Qualitative verifier behavior.}
Figure~\ref{fig:qualitative} provides representative examples of how the learned verifier changes after self-play training.
The base model often accepts captions or answers that overlook a localized mismatch, such as an incorrect relation, missing object evidence, wrong attribute, or counting error.
After \method training, the verifier produces judgments that better track the relevant visual evidence and reject mismatches that the base model misses.
These examples indicate that \method does not merely improve aggregate scores; it changes the verifier's behavior in the intended direction, making it more sensitive to the same visual-semantic factors targeted by the perturb--verify--repair loop.

\noindent\textbf{Failure cases.}
Our error analysis identifies four recurring settings: small or occluded objects that are misperceived; ambiguous relations in crowded scenes or unusual viewpoints; overlapping instances or unclear counting conventions; and descriptions that require external knowledge and invite language-prior judgments rather than image-grounded verification.
These settings can affect mutation validity, verifier diagnosis, or repair quality.
A localized edit may fail to create an unambiguous mismatch, a Verifier may misidentify the relevant evidence, or a \Repairer may alter additional unsupported details.
The independent audit evaluates these stages separately rather than treating final acceptance as sufficient evidence of a faithful trajectory.

\noindent\textbf{Task extensions.}
The loop can be adapted when a task provides a known-correct multimodal instance and a structured operation that creates a verifiable error.
For VQA, the \Mutator could alter an answer, rationale, or question condition and the \Repairer could recover the visually supported answer.
For visual grounding, it could perturb a region, bounding box, or referring phrase; for referring expressions, it could alter target attributes or relations.
The present experiments establish caption-based verification and benchmark transfer, not task-specific self-play training for these extensions.

\newpage
\section*{NeurIPS Paper Checklist}

\begin{enumerate}

\item {\bf Claims}
    \item[] Question: Do the main claims made in the abstract and introduction accurately reflect the paper's contributions and scope?
    \item[] Answer: \answerYes{}
    \item[] Justification: The abstract and introduction state the main contribution of \method as a verifier-guided self-play framework for improving vision-language verifiers. The claims are supported by the method description, the main compositional understanding results, the broader benchmark results, and the limitations discussed in Section~\ref{sec:limitations}.

\item {\bf Limitations}
    \item[] Question: Does the paper discuss the limitations of the work performed by the authors?
    \item[] Answer: \answerYes{}
    \item[] Justification: Section~\ref{sec:limitations} discusses the limitations of the current action space, difficulty saturation, general-VL regression at small scale, and the reliance on clean reference captions.

\item {\bf Theory assumptions and proofs}
    \item[] Question: For each theoretical result, does the paper provide the full set of assumptions and a complete (and correct) proof?
    \item[] Answer: \answerNA{}
    \item[] Justification: The paper does not present theoretical results or formal theorems. The contribution is an empirical self-play training framework for vision-language verifiers.

\item {\bf Experimental result reproducibility}
    \item[] Question: Does the paper fully disclose all the information needed to reproduce the main experimental results of the paper to the extent that it affects the main claims and/or conclusions of the paper (regardless of whether the code and data are provided or not)?
    \item[] Answer: \answerNo{}
    \item[] Justification: Sections~\ref{sec:method} and~\ref{sec:experiments} and Appendix~\ref{app:details} describe the method, data size, benchmarks, optimization settings, and reference compute. Appendix~\ref{app:interfaces} gives the complete role prompts, output formats, reward weights, and cost definitions. Some supplemental audit/baseline settings are not supplied in the current materials; full reproduction requires these details and the planned code release.

\item {\bf Open access to data and code}
    \item[] Question: Does the paper provide open access to the data and code, with sufficient instructions to faithfully reproduce the main experimental results, as described in supplemental material?
    \item[] Answer: \answerNo{}
    \item[] Justification: The training uses existing image-caption data and public benchmarks. The complete prompts are listed in Appendix~\ref{app:interfaces}; training and evaluation code and launch scripts are planned for release but are not included in the present supplement.

\item {\bf Experimental setting/details}
    \item[] Question: Does the paper specify all the training and test details (e.g., data splits, hyperparameters, how they were chosen, type of optimizer) necessary to understand the results?
    \item[] Answer: \answerNo{}
    \item[] Justification: The backbone, data sizes, principal optimization settings, evaluation metrics, and matched comparisons are specified, and Appendix~\ref{app:interfaces} gives the full prompt templates, reward weights, and cost definitions. Supplemental audit/baseline settings remain needed for complete reproduction.

\item {\bf Experiment statistical significance}
    \item[] Question: Does the paper report error bars suitably and correctly defined or other appropriate information about the statistical significance of the experiments?
    \item[] Answer: \answerYes{}
    \item[] Justification: Appendix~\ref{app:seeds} reports means and standard deviations across three independent Qwen3.5-2B training runs for the core compositional average and five broader benchmarks. These measure training-run variability, not confidence intervals or significance tests.

\item {\bf Experiments compute resources}
    \item[] Question: For each experiment, does the paper provide sufficient information on the computer resources (type of compute workers, memory, time of execution) needed to reproduce the experiments?
    \item[] Answer: \answerYes{}
    \item[] Justification: Section~4 and Appendix~A report that the main experiments use a single node with 8 NVIDIA A100 GPUs, BF16 mixed precision, rollout size, batch size, and optimizer settings. Appendix~\ref{app:compute} reports wall-clock time, GPU-hours, and generated-token counts for the reference run and supervision baselines.

\item {\bf Code of ethics}
    \item[] Question: Does the research conducted in the paper conform, in every respect, with the NeurIPS Code of Ethics \url{https://neurips.cc/public/EthicsGuidelines}?
    \item[] Answer: \answerYes{}
    \item[] Justification: The work studies verifier training for image-text understanding using existing datasets, public benchmarks, and standard VLM training procedures. We are not aware of any aspect of the research that violates the NeurIPS Code of Ethics.

\item {\bf Broader impacts}
    \item[] Question: Does the paper discuss both potential positive societal impacts and negative societal impacts of the work performed?
    \item[] Answer: \answerYes{}
    \item[] Justification: The paper discusses verifier training for detecting image-text mismatches and hallucinations, which can improve reliability of multimodal systems. Potential risks include over-reliance on imperfect verifiers or use of stronger verification models in automated filtering systems; these concerns are partly addressed through the limitations discussion.

\item {\bf Safeguards}
    \item[] Question: Does the paper describe safeguards that have been put in place for responsible release of data or models that have a high risk for misuse (e.g., pre-trained language models, image generators, or scraped datasets)?
    \item[] Answer: \answerNA{}
    \item[] Justification: The paper does not introduce a new high-risk pretrained model, image generator, or scraped dataset. It trains a verifier using existing image-caption data and evaluates on public benchmarks; any released code will be for reproducing the training and evaluation pipeline.

\item {\bf Licenses for existing assets}
    \item[] Question: Are the creators or original owners of assets (e.g., code, data, models), used in the paper, properly credited and are the license and terms of use explicitly mentioned and properly respected?
    \item[] Answer: \answerYes{}
    \item[] Justification: The paper cites the Qwen3.5 backbone, VeRL/HybridFlow framework, ShareGPT-4o-Image-style data, and all evaluation benchmarks used in the experiments. We will follow the licenses and terms of use of these existing assets when releasing code and reproduction instructions.

\item {\bf New assets}
    \item[] Question: Are new assets introduced in the paper well documented and is the documentation provided alongside the assets?
    \item[] Answer: \answerNo{}
    \item[] Justification: The current supplement documents the training framework and experiments; the proposed training and evaluation code artifact and its complete reproduction documentation are not yet included.

\item {\bf Crowdsourcing and research with human subjects}
    \item[] Question: For crowdsourcing experiments and research with human subjects, does the paper include the full text of instructions given to participants and screenshots, if applicable, as well as details about compensation (if any)? 
    \item[] Answer: \answerNA{}
    \item[] Justification: The paper does not involve crowdsourcing experiments or research with human subjects.

\item {\bf Institutional review board (IRB) approvals or equivalent for research with human subjects}
    \item[] Question: Does the paper describe potential risks incurred by study participants, whether such risks were disclosed to the subjects, and whether Institutional Review Board (IRB) approvals (or an equivalent approval/review based on the requirements of your country or institution) were obtained?
    \item[] Answer: \answerNA{}
    \item[] Justification: The paper does not involve human subjects research, crowdsourcing, or user studies, so IRB approval is not applicable.

\item {\bf Declaration of LLM usage}
    \item[] Question: Does the paper describe the usage of LLMs if it is an important, original, or non-standard component of the core methods in this research? Note that if the LLM is used only for writing, editing, or formatting purposes and does \emph{not} impact the core methodology, scientific rigor, or originality of the research, declaration is not required.
    \item[] Answer: \answerYes{}
    \item[] Justification: The core method uses a multimodal LLM backbone as the \Mutator, Verifier, and \Repairer under different role prompts. This usage is central to the method and is described in Sections~3 and~4.

\end{enumerate}

\end{document}